\documentclass[letterpaper]{article} 
\usepackage{aaai25}  
\usepackage{times}  
\usepackage{helvet}  
\usepackage{courier}  
\usepackage[hyphens]{url}  
\usepackage{graphicx} 
\usepackage{natbib}  
\usepackage{caption} 
\usepackage{algorithm}
\usepackage{algorithmic}

\usepackage{newfloat}
\usepackage{listings}
\DeclareCaptionStyle{ruled}{labelfont=normalfont,labelsep=colon,strut=off} 
\floatstyle{ruled}
\newfloat{listing}{tb}{lst}{}
\floatname{listing}{Listing}
\title{\our{}: Explainability of Image Classification Models by Information Disentanglement}
\author {
    \L{}ukasz Struski$^{\,1}$,\; Dawid Rymarczyk$^{\,1, 2}$,\; Jacek Tabor$^{\,1}$
}
\affiliations {
    $^{1}$ Faculty of Mathematics and Computer Science \\ 
    Jagiellonian University, Krak\'ow, Poland \\[0.3em]
    $^{2}$ Ardigen SA\\[0.3em]
    {\small \{lukasz.struski, dawid.rymarczyk, jacek.tabor\}@uj.edu.pl}
}

\usepackage{bibentry}

\usepackage{amsmath}
\usepackage{amssymb}
\usepackage{cleveref}
\usepackage{booktabs}
\usepackage{multirow}
\usepackage{array, makecell} %

\usepackage{arydshln}

\usepackage{subcaption}

\makeatletter
\def\adl@drawiv#1#2#3{%
        \hskip.5\tabcolsep
        \xleaders#3{#2.5\@tempdimb #1{1}#2.5\@tempdimb}%
                #2\z@ plus1fil minus1fil\relax
        \hskip.5\tabcolsep}
\newcommand{\cdashlinelr}[1]{%
  \noalign{\vskip\aboverulesep
           \global\let\@dashdrawstore\adl@draw
           \global\let\adl@draw\adl@drawiv}
  \cdashline{#1}
  \noalign{\global\let\adl@draw\@dashdrawstore
           \vskip\belowrulesep}}
\makeatother

\DeclareMathOperator*{\argmax}{arg\,max}

\def\R{\mathbb{R}}

\def\mxpool{\mathrm{mx\_pool}}

\def\relu{\mathrm{ReLU}}

\def\our#1{InfoDisent#1}

\usepackage{color}

\usepackage{cuted}

\begin{document}

\maketitle

\begin{abstract}

In this work, we introduce InfoDisent, a hybrid approach to explainability based on the information bottleneck principle. InfoDisent enables the disentanglement of information in the final layer of any pretrained model into atomic concepts, which can be interpreted as prototypical parts. This approach merges the flexibility of post-hoc methods with the concept-level modeling capabilities of self-explainable neural networks, such as ProtoPNets. We demonstrate the effectiveness of InfoDisent through computational experiments and user studies across various datasets using modern backbones such as ViTs and convolutional networks. Notably, InfoDisent generalizes the prototypical parts approach to novel domains (ImageNet).

\end{abstract}


\section{Introduction}
\label{sec:intro}


Deep neural networks have demonstrated performance that matches or even surpasses human capabilities across various domains, such as image classification and generation, speech recognition, and natural language processing. Despite their impressive achievements, these networks often operate as "black boxes", offering little insight into the reasoning behind their decisions~\cite{rudin2019stop}. 

This lack of transparency poses significant challenges, particularly in high-stakes applications such as medicine and autonomous driving, where understanding a model's decision-making process is critical~\cite{bojarski2017explaining,khan2001classification,nauta2023anecdotal,patricio2023explainable,samek2021explaining,struski2024multiple}. To address this issue, the subfield of artificial intelligence known as eXplainable AI (XAI)~\cite{xu2019explainable} has emerged, focusing on making AI systems more interpretable and trustworthy.


XAI methods can be classified into two major categories: post-hoc methods and inherently interpretable (ante-hoc) methods~\cite{rudin2019stop}. Post-hoc methods, such as GradCAM~\cite{selvaraju2017grad}, are highly flexible because they can be applied to any neural network architecture. However, they are often unreliable~\cite{adebayo2018sanity}, provide local explanations, and fail to reveal the characteristics of specific classes. In contrast, ante-hoc methods, such as ProtoPNet~\cite{chen2019looks}, offer concept-level explanations by identifying prototypical parts from the training dataset. While this approach enhances interpretability, it is
\begin{figure}[t!]
    \centering
    \includegraphics[width=0.99\linewidth]{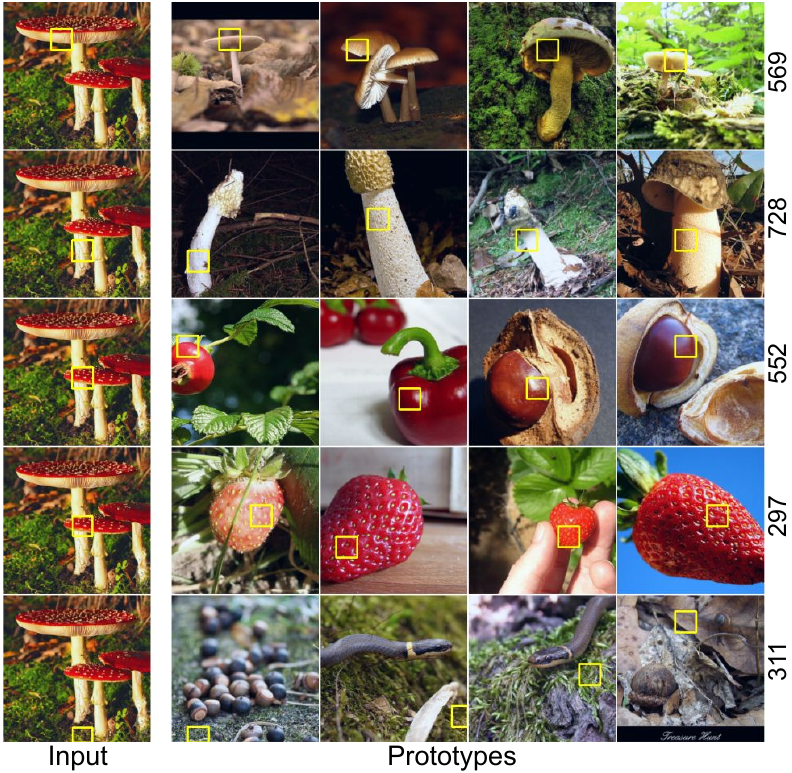}
    \caption{Decision explanation constructed by \our{} for the pre-trained ViT feature space on the \textit{Agaric} mushrooms image from the ImageNet. We can trace the decision of ViT behind assigning the class \textit{Agaric} to the image on the left to having a hat (569), a white leg (728), a reddish shine (552), a strawberry texture (297) and the appearance of ground with moss (311). Note that, the prototype block (right) each row represents the prototypical part (the corresponding channel number). The yellow boxes in each row show the activation of a given prototypical part, while in the first column, we show the activation of corresponding prototypical parts in the original image. 
    }
    \label{fig:teaser}
\end{figure}
limited to fine-grained classification tasks and specific architectures, such as CNNs in the case of ProtoPNet. Consequently, these methods do not fully leverage the potential of large pretrained models, such as Vision Transformers (ViTs)~\cite{dosovitskiy2020image}.

To address this, we propose \our{}, a novel XAI approach that leverages information disentanglement in the final layer of any pretrained model. In \our{}, each channel encapsulates a single atomic concept, interpretable as prototypical parts -- similar to the approach used in PIPNet~\cite{nauta2023pipnet}. This is achieved through the application of an information bottleneck, in which we enforce activation sparsity in each prototypical channel. As a result, \our{} provides both local and global-level explanations in the form of human-interpretable concepts, as illustrated in~\Cref{fig:teaser}. Furthermore, \our{} offers significant flexibility, as it can be seamlessly applied to any pretrained architecture, including ViTs and convolutional networks.

We demonstrate the effectiveness of \our{} through both computational experiments and user studies. Our method is benchmarked on five datasets, including ImageNet, a challenging dataset where prototypical parts-based approaches have not been generalized to. Additionally, results from the user study indicate that \our{} provides a competitive level of explanation understanding compared to other prototypical parts-based methods, while offering the advantage of flexibility in its application to any model backbone. We make the code available.

Our contributions can be summarized as follows: \begin{itemize} 
\setlength{\itemsep}{1pt} 
\setlength{\parskip}{0pt} 
\setlength{\parsep}{0pt} 
\item We propose \our{}, a hybrid XAI model that combines the interpretability of ante-hoc methods with the flexibility of post-hoc approaches. 
\item To the best of our knowledge, \our{} is the first model to provide prototypical, part-like interpretations for the feature space of any pretrained network. 
\item We validate the effectiveness of \our{} in terms of both accuracy and user understanding through extensive experimental evaluations.
\end{itemize}

\section{Related Works}

Research in XAI can be divided into two disjoint categories: post-hoc interpretability \cite{lundberg2017unified,ribeiro2016should,selvaraju2017grad}, where we analyze the pre-trained model to explain its predictions, and inherently explained models~\cite{bohle2022b,chen2019looks}, where the aim lies in building networks which decisions are easy to interpret. Both of the above approaches have their advantages and disadvantages which we discuss in the following paragraphs.

\paragraph{Post-hoc methods} In the post-hoc methods, we interpret existing pre-trained network architectures. The commonly used methods such as SHAP~\cite{lundberg2017unified,shapley1951notes}, LIME~\cite{ribeiro2016should}, LRP~\cite{bach2015pixel} and Grad-CAM~\cite{selvaraju2017grad} provide in practice only feature importance which can be visualized as a saliency map that shows on which part of the image the model has focused its attention. 
This allows us to check if the model does not focus its attention outside of the object of interest \cite{ribeiro2016should}, however, it is in general not sufficient to really understand the reasons behind given predictions. Additionally, post-hoc methods allow typically only local explanations (per the prediction of a given image), and do not allow to understand of the prerequisites to the given class.

\paragraph{Inherently explained models}
While post-hoc methods are easy to implement due to their non-intrusive nature, they often produce biased and unreliable explanations~\cite{NEURIPS2018_294a8ed2}. To address this, recent research has increasingly focused on designing self-explainable models that make the decision process directly visible~\cite{brendel2018approximating,NEURIPS2018_3e9f0fc9}. Many of these interpretable solutions utilize attention mechanisms~\cite{liu2021visual,zheng2019looking} or exploit the activation space, such as with adversarial autoencoders~\cite{Guidotti_Monreale_Matwin_Pedreschi_2020}. Among the most recent approaches, ProtoPNet~\cite{chen2019looks} has significantly influenced the development of self-explainable models.
It learns class-specific prototypes, similar to concepts, with a fixed number per class. The model classifies inputs by calculating responses from each class’s prototypes and summarizing these responses through a fully connected layer, providing explanations as a weighted sum of all prototypes. This method inspired the development of several other self-explainable models~\cite{donnelly2022deformable,nauta2023pipnet,pach2024lucidppn,rymarczyk2021protopshare,rymarczyk2022interpretable,rymarczyk2023icicle,wang2021interpretable}. 
Typically, in the prototypical parts-based model, the final decision of a given class is decomposed into the appearance of a few selected prototypes, which are similar to some strongly localized parts of some chosen images from the training dataset. 
We see that in this approach the final class decision is split into a set of simpler and interpretable concepts. 


\begin{figure*}[thb]
    \centering
    \includegraphics[width=\linewidth]{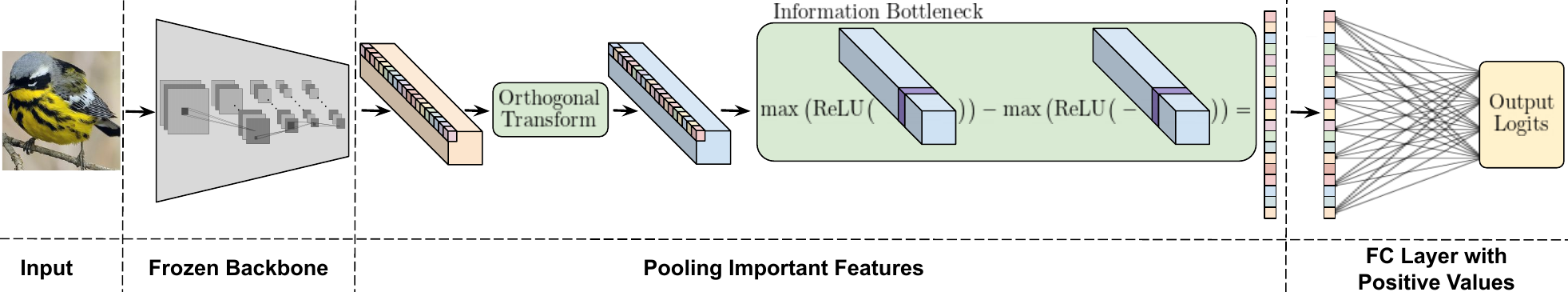}
    \caption{
    Our image classification interpretation model, \our{}, features three main components: a pre-trained backbone, a pooling layer for key features, and a fully connected layer. The CNN/transformer backbone, with frozen weights, is not further trained. The pooling layer extracts features from the last transformer or convolutional layer and identifies key positive and negative features. These are then combined into a dense vector, which is processed by a fully connected linear layer with nonnegative entries in the final stage.}
    \label{fig:architecture}
\end{figure*}

\section{\our{}}
\label{sec:model}

Our motivation comes from the general methodology of prototypical models, which can be described as follows:
\begin{itemize}
  \setlength{\itemsep}{1pt}
  \setlength{\parskip}{0pt}
  \setlength{\parsep}{0pt}
    \item trace the final class decision to the co-occurrence in the image of some prototypes,
    \item each prototype can be visualized by some prototypes from the training dataset,
    \item prototypes should be strongly localized parts of the image.
\end{itemize}

We claim that even for pre-trained models it is possible to disentangle the channels in the feature space in such a way that they will become sufficiently informative to satisfy the above points. In \our{} we disentangle the channels in feature space by applying an orthogonal map in the pixel space, and consequently do not change the inner-lying distance and scalar product.

Since we disentangle the channels in the feature space, we restrict our attention to images in the feature space and consider only the classification head.
Thus consider the data set of images in the feature space, which consists of images of possibly different resolutions but the same number of channels $d$. 

\paragraph{Default classification head in deep networks}
To establish notation, let us first describe the typical classification head. In the case of a classification task with $k$ classes, we apply the following operations for the image $I$ in the feature space:
\begin{enumerate}
    \item $I \to v_I=\mathrm{avg\_pool\_over\_channels}(I) \in \R^d$ 
    \item $v_I \to w_I=Av_I$, where $A$ is 
 a matrix of dimensions $d \times k$
 \item $w_I \to p_I=\mathrm{softmax}(w_I)$.
\end{enumerate}

In the case of \our{} we apply a few mechanisms to ensure the desired properties. First, we need to be able to disentangle the channel space. To do so we apply the unitary map $U$ in the pixel space. Next, we apply the information bottleneck -- for a given channel instead of the average pool where all pixels participate, we use extremely sparse analog where only the value of the highest positive and negative pixels are used. 
Finally, as is common in interpretable methods we use matrix $A$, but only with nonnegative coefficients to allow only positive reasoning.

\paragraph{The sparse pooling features mechanism} Most interpretable models involve retraining certain parts of the CNN~\cite{chen2019looks,rymarczyk2022interpretable}, whereas others, like PIP-Net~\cite{nauta2023pipnet}, retrain the entire CNN. In contrast, our method utilizes a pre-trained CNN or transformer without further modification during training. We first employ a trainable orthogonal transformation $U$ on pixel space to enable the disentanglement of hidden features from feature maps\footnote{To parametrize orthogonal maps, we restrict to those with positive determinants and use the formula
$U=\exp(W-W^T)$, where $\exp(\cdot)$ denote the matrix exponential~\cite{hall2013lie}. We utilize the fact that the space of orthogonal matrices with positive determinants coincide with exponentials of skew-symmetric matrices
\cite{shepard2015representation}. 
}. To enforce the disentanglement we follow by an introduced sparse analogue of average pooling over channel $K$ given by 
$$
K \to \mxpool(K)=\max(\relu(K))-\max(\relu(-K)).
$$
Observe that to compute $\mxpool(K)$ we need to know only the highest positive and negative pixel values of $K$, contrary to $\mathrm{avg\_pool}$, where
all the pixel values are needed. 
Subsequently, we identify sparse representations (superpixels) within the channels that contribute positively or negatively to predictions. This enables us to generate heatmaps akin to Grad-CAM~\cite{selvaraju2017grad} without necessitating a backward model step, as shown in~\Cref{fig:heatmap_explanations}. Importantly, unlike Grad-CAM, our technique supports the visualization of negative heatmaps, resembling the LRP~\cite{bach2015pixel} method that requires a backward pass in a neural network. Our method operates solely during the forward step. 

Finally, as is common in XAI models, to allow only positive reasoning we allow the matrix $A$ to have only nonnegative values.





\paragraph{Classification head in \our{}}
Finally, the classification head in \our{} is given by:
\begin{enumerate}
    \item $I=(I_{rs})_{rs} \to J=(UI_{rs})$, where $U:\R^d \to \R^d$ is an orthogonal matrix and $I_{rs}$ denotes the pixel value of $I$ with coordinates $r$ and $s$,
    \item $J \to v_J=\mathrm{mx\_pool\_over\_channels}(J) \in \R^d$, where for a given channel $K$ we have
    $$
    \mathrm{mx\_pool}(K)=\max(\relu(K))-\max(\relu(-K)),
    $$
    \item $v_J \to w_J=Av_J$, where $A$ is 
 a matrix with nonnegative coefficients of dimensions $d \times k$
 \item $w_J \to p_I=\mathrm{softmax}(w_J)$
\end{enumerate}

\paragraph{\our{} model}
Thus \our{} consists of two main components: the frozen CNN or transformer Backbone, and \our{} classification head, as illustrated in~\Cref{fig:architecture}.

The first component, the CNN or transformer Backbone, is a frozen classical pre-trained network up to the final feature maps layer. 
Importantly, this part of the network and its weights remain unaltered during the training of \our{}, ensuring that the learned feature representations are preserved.

The second component involves pooling important features from the final feature maps produced by the Frozen Backbone. This pooling mechanism encourages the network to use information bottlenecks to build maximally informative channels independent of each other. This leads to constructing prototypical channels, which can be easily interpreted. 
The final element is the fully connected layer, which is distinguished by its positive weight values. This restriction on positivity, except biases handled as in conventional linear layers, ensures that the information about the positive or negative contributions of selected features from the previous part of the model is preserved. This constraint is beneficial for the interpretability of the model's predictions, as it clarifies the contribution of each feature to the final output.

Observe, that contrary to some Post-hoc methods \our{} need training of the model on the whole dataset.

\section{Understanding the classification decisions}

\begin{figure}[t]
    \centering
    \includegraphics[width=\linewidth]{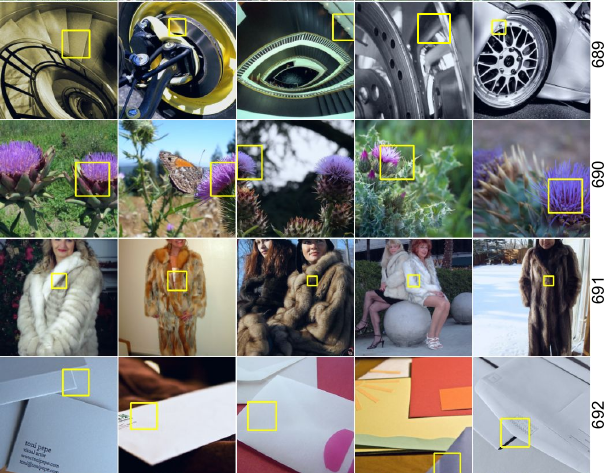}
    \caption{The image shows prototypes from channels 689 to 692 in a trained ResNet-50 on the ImageNet. Each row displays the 5 most significant patches from a single prototypical channel. The prototype's activations are highlighted by yellow boxes.}
    \label{fig:prototypes_channels}
\end{figure}

\paragraph{Prototypes in  \our{}}
The crucial consequence of \our{} that is disentangles the channels making them interpretable. Thus, similarly to PiPNet \cite{nauta2023pipnet}, we identify channels as prototypes. To illustrate a given prototype channel, we present five images from the training dataset on which the activation of the channel is the greatest, see~\Cref{fig:prototypes_channels}, where we present consecutive prototypes for the pre-trained ResNet-50 model. This follows from the fact that for better interpretability of model decisions, it is beneficial for humans to be presented from 4 to 9 concepts~\cite{rymarczyk2022interpretable}.
Formally, similarly as in prototypical models,  
as the prototypical part, we understand the part of the image corresponding to pixels in feature space with maximal activation, marked by the yellow box in ~\Cref{fig:prototypes_channels}. 
Observe that the presented prototypes seem consistent with each other, and could be well interpreted.

\paragraph{Understanding the Decision for a Given Image by Prototypes}

Now that we can understand and visualize prototypes, there appears to be a question of how to visualize the prototypes crucial for the decision of the model on the given image. To do this we chose 5
prototypical channels which are most important for the prediction\footnote{One can assume more subtle strategies, see Supplementary Materials.} in~\Cref{fig:prototypes_image}.
For each channel, we identified 5 images from the training dataset that exhibit the strongest activation values for that channel, as depicted by the red spots in~\Cref{fig:heatmap_explanations}. In simpler terms, we selected the top 5 images based on the highest activation values, or $\mathrm{arg}$-$\mathrm{top5}$, for each channel.

The model's proposed prototypes are easy to interpret. Moreover, unlike current state-of-the-art prototype methods, our model excels at interpreting images from the ImageNet dataset. More examples from various datasets and models are presented in the Supplementary Materials.

\begin{figure}[thb]
    \centering
    \includegraphics[width=\linewidth]{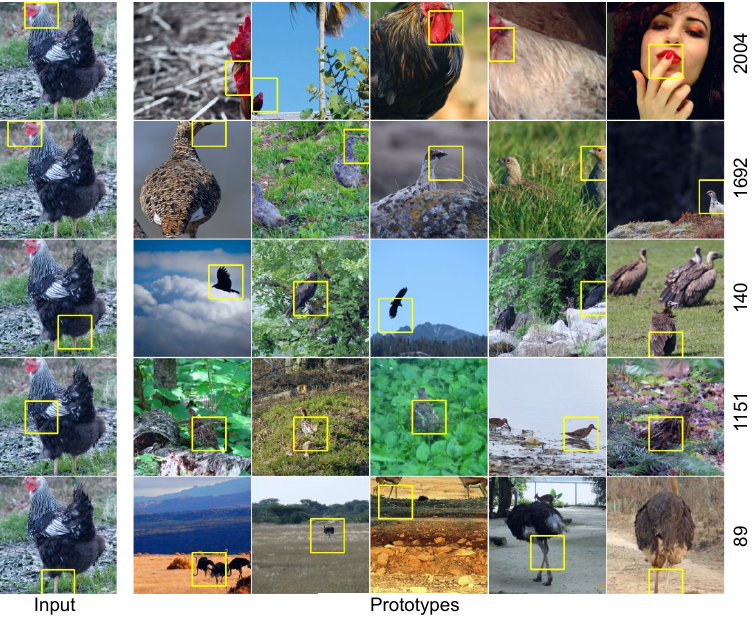}
    \caption{Exemplary explanation (\textit{hen}) for ResNet-50 backbone provided by \our{} in a form of prototypical parts.}
    \label{fig:prototypes_image}
\end{figure}


\begin{figure}[thb]
  \centering
  \begin{subfigure}{\linewidth}
    \includegraphics[width=\linewidth]{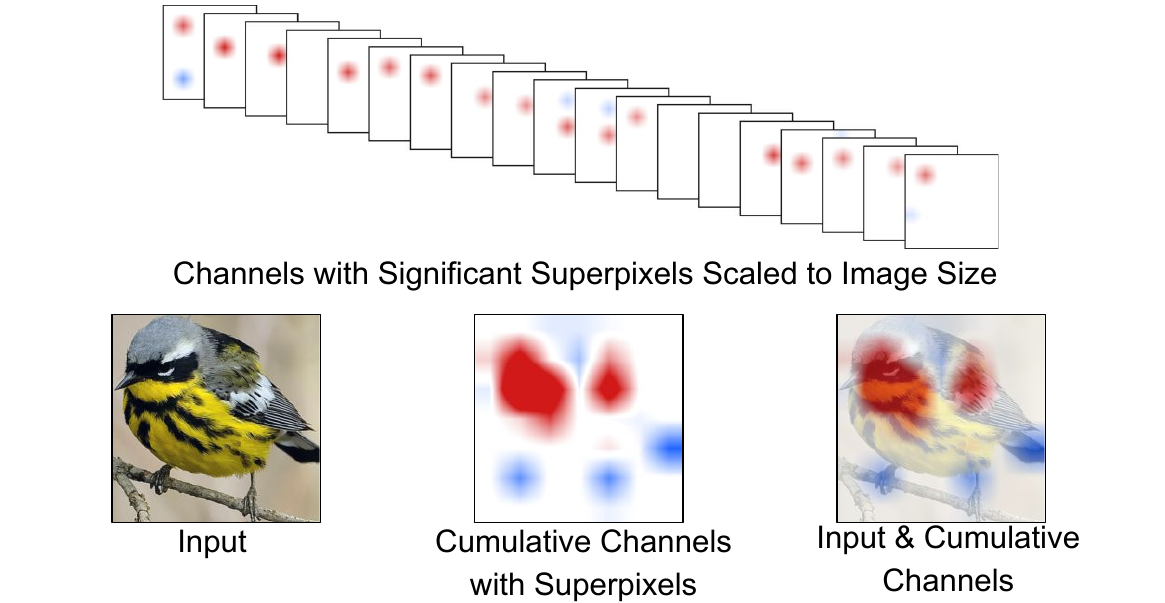}
    \caption{The top part shows a heatmap with positive (red) and negative (blue) sparse representations, each channel displaying one positive and one negative superpixel. These sparse representations are then aggregated and rescaled to the image size in the lower part.}
    \label{fig:heatmap_explanations}
  \end{subfigure}
  \\[0.5em]
  \begin{subfigure}{\linewidth}
    \includegraphics[width=\linewidth]{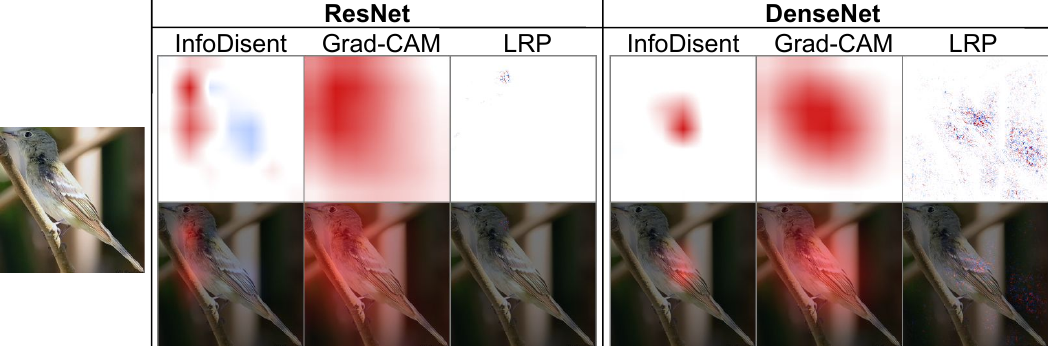}
    \caption{Example compares visual explanations from \our{}, Grad-CAM, and LRP. The left shows input images, with the following columns displaying explanations from these methods applied to ResNet and DenseNet.}
    \label{fig:heatmaps}
  \end{subfigure}
  \caption{The image demonstrates how to analyze and visualize decisions made by \our{}.}
  \label{fig:heatmap_solution}
\end{figure}

\paragraph{Heatmaps}
Our approach, which relies on representation channels, enables us to easily generate heatmaps similar to those produced by the Grad-CAM method, see \Cref{fig:heatmap_explanations}. To do this we accumulate the activations of all prototypes (both positive and negative ones) over all channels. Since in \our{} we use information bottleneck, we obtain more localized results than other standard approaches.

Observe that, unlike Grad-CAM, our heatmaps also illustrate negative activations, akin to the Layer-Wise Relevance Propagation (LRP) method. While LRP can effectively highlight both positive and negative contributions to the model's decision, it often faces challenges such as complexity in implementation and sensitivity to model architecture variations. Our method addresses these issues by providing clear and interpretable visualizations. Sample results using this explanation method are presented in~\Cref{fig:heatmaps}.


\begin{figure}[thb]
    \centering
    \includegraphics[width=1\linewidth]{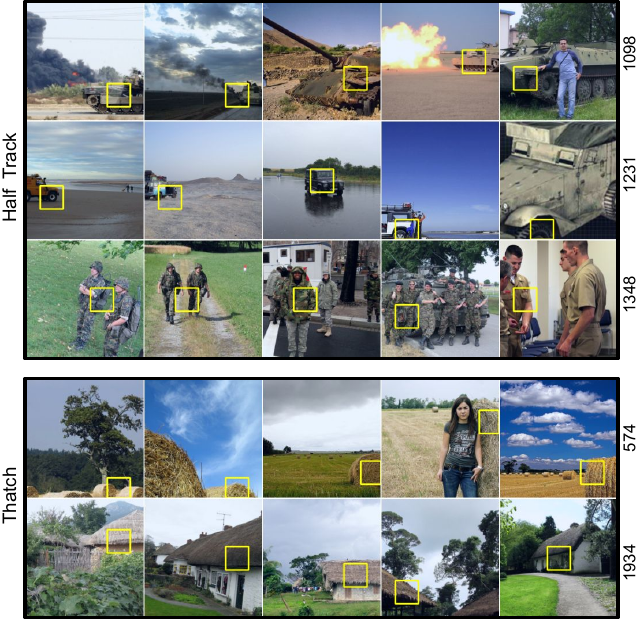}
    \caption{Prototypes of two classes (from top to bottom): Half Track and Thatch from the ImageNet dataset. Each row displays the 5 most significant prototypes for a specific channel, with the channel numbers listed on the right. The order of the prototypes within each row reflects their importance for explaining the given class. In the Half Track class, the prototypes from channel 1098 are the most crucial, followed by those from channels 1231 and 1348. These prototypes effectively explain the Half Track class, as they highlight elements of a tank (from channel 1098), a car, and the possible presence of soldiers on board.}
    \label{fig:prototypes_class}
\end{figure}

\paragraph{Understanding the Decision Behind Class} 

We examine the model's decision-making process on a per-class basis by utilizing prototypes. To identify prototypes for a given class, we focus on key channels that are prominently activated across all test set images belonging to that class. These key channels are selected based on their consistent presence and strong activation in images of the same class. Once we have identified these crucial channels, we visualize the prototypes as previously described, providing a clear representation of what the model deems important for that particular class.

\Cref{fig:prototypes_class} illustrates the prototypes for selected classes from the ImageNet dataset. Each prototype captures essential features such as material types, structural elements, or specific textures that are characteristic of the class. 

\begin{table}[!t]  
  \centering
  \small
  \caption{Accuracy comparison of interpretability models using standard CNN architectures (utilized in explainable models) trained on cropped bird images of CUB-200-2011, and Stanford Cars (Cars). Our approach demonstrates superior performance across nearly all the datasets and models considered. For each dataset and backbone, we boldface the best result in the class of interpretable models.}
  \begin{tabular}{@{\;\,}l@{\;\,}l@{\quad}c@{\quad}c@{\;\,}}
    \toprule
    & \multirow{2}{*}{\textbf{Model}} & \multicolumn{2}{c}{\textbf{Dataset}}  \\
    \cmidrule(l{0pt}r{10pt}){3-4}
    && CUB-200-2011 & Cars \\
    \midrule
    \multirow{7}{*}{\rotatebox{90}{\footnotesize \textit{ResNet-34}}} & ResNet-34 & 82.4\% & 92.6\% \\
    & \hspace{5pt}\includegraphics[width=6pt]{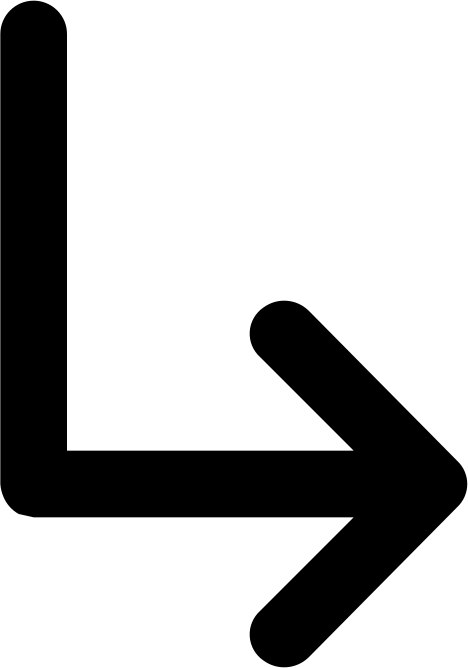}\,\our{\,(ours)} & \textbf{83.5}\% & \textbf{92.8}\% \\
    \cdashlinelr{2-4}
    & ProtoPNet & 79.2\% & 86.1\%  \\
    & ProtoPShare & 74.7\% & 86.4\%  \\
    & ProtoPool & 80.3\% & 89.3\%  \\
    & ST-ProtoPNet & \textbf{83.5}\% & 91.4\%  \\
    & TesNet & 82.7\% & 90.9\%  \\
    \cmidrule(lr){1-4}
    \multirow{5}{*}{\rotatebox{90}{\footnotesize \textit{ResNet-50}}} & ResNet-50 & 83.2\% & 93.1\% \\
    & \hspace{5pt}\includegraphics[width=6pt]{images/subdirectory_arrow_right.png}\,\our{\,(ours)} & \textbf{83.0}\% & \textbf{92.9}\% \\
    \cdashlinelr{2-4}
    & ProtoPool & -- & 88.9\%  \\
    & ProtoTree & -- & 86.6\%  \\
    & PIP-Net & 82.0\% & 86.5\%  \\
    \cmidrule(lr){1-4}
    \multirow{7}{*}{\rotatebox{90}{\footnotesize \textit{DenseNet-121}}} & DenseNet-121 & 81.8\% & 92.1\% \\
    & \hspace{5pt}\includegraphics[width=6pt]{images/subdirectory_arrow_right.png}\,\our{\,(ours)} & 82.6\% & \textbf{92.7}\% \\
    \cdashlinelr{2-4}
    & ProtoPNet & 79.2\% & 86.8\%  \\
    & ProtoPShare & 74.7\% & 84.8\%  \\
    & ProtoPool & 73.6\% & 86.4\%  \\
    & ST-ProtoPNet & \textbf{85.4}\% & 92.3\%  \\
    & TesNet & 84.8\% & 92.0\%  \\
    \cmidrule(lr){1-4}
    \multirow{3}{*}{\rotatebox{90}{\footnotesize \textit{ConvNeXt}}} & ConvNeXt-Tiny & 83.8\% & 91.0\% \\
    & \hspace{5pt}\includegraphics[width=6pt]{images/subdirectory_arrow_right.png}\,\our{\,(ours)} & 84.1\% & \textbf{90.2}\% \\
    \cdashlinelr{2-4}
    & PIP-Net & \textbf{84.3}\% & 88.2\%  \\
    \bottomrule
  \end{tabular}
  
  \label{tab:acc_cropped_data}
\end{table}

\section{Numerical Experiments}
\label{sec:ex_results}

This section outlines the experiments conducted to compare our approach with current state-of-the-art methods, highlighting the most significant results. 
Our experiments utilized a variety of datasets, including well-established benchmarks for evaluating the explainability of artificial models: The Caltech-UCSD Birds-200-2011 (CUB-200-2011)~\cite{wah2011caltech}, Stanford Cars~\cite{krause20133d}, and Stanford Dogs~\cite{khosla2011novel}. Additionally, we incorporated the full ImageNet~\cite{russakovsky2015imagenet} dataset for the first time in the class of prototypical networks. 

Additional results, including ablation study and analysis of inter-channel correlation, and details concerning those experiments are provided in the Supplementary Materials.


\begin{table}[t]  
  \centering
  \caption{Classification accuracy on full CUB-200-2011, and Stanford Dogs datasets by competing approaches using different CNN backbones. 
  For each dataset and backbone, we boldface the best result in the class of interpretable models.}
  \begin{tabular}{@{\;\,}l@{\;\,}l@{}c@{\quad}c@{\;\,}}
    \toprule
    & \multirow{2}{*}{\textbf{Model}} & \multicolumn{2}{c}{\textbf{Dataset}} \\
    \cmidrule(r){3-4}
    && CUB-200-2011 & Dogs  \\
    \midrule
    \multirow{5}{*}{\rotatebox{90}{\footnotesize \textit{ResNet-34}}} & ResNet-34 & 76.0\% & 84.5\% \\
    & \hspace{5pt}\includegraphics[width=6pt]{images/subdirectory_arrow_right.png}\,\our{\,(ours)} & \textbf{78.3}\% & \textbf{83.9}\% \\
    \cdashlinelr{2-4}
    & ProtoPNet & 74.1\% & 76.1\% \\
    & ST-ProtoPNet & 78.2\% & 83.4\% \\
    & TesNet & 76.5\% & 81.2\% \\
    \cmidrule(lr){1-4}
    \multirow{5}{*}{\rotatebox{90}{\footnotesize \textit{ResNet-50}}} & ResNet-50 & 78.7\% & 87.4\% \\
    & \hspace{5pt}\includegraphics[width=6pt]{images/subdirectory_arrow_right.png}\,\our{\,(ours)} & 79.5\% & \textbf{86.6}\% \\
    \cdashlinelr{2-4}
    & ProtoPNet & 84.8\% & 78.1\% \\
    & ST-ProtoPNet & \textbf{88.0}\% & 83.3\%  \\
    & TesNet & 87.3\% & 85.7\%  \\
    \cmidrule(lr){1-4}
    \multirow{5}{*}{\rotatebox{90}{\footnotesize \textit{DenseNet-121}}} & DenseNet-121 & 78.2\% & 84.1\% \\
    & \hspace{5pt}\includegraphics[width=6pt]{images/subdirectory_arrow_right.png}\,\our{\,(ours)} & 80.6\% & \textbf{83.8}\%  \\
    \cdashlinelr{2-4}
    & ProtoPNet & 76.6\% & 75.4\%  \\
    & ST-ProtoPNet & \textbf{81.8}\% & 82.9\%  \\
    & TesNet & 80.9\% & 82.1\%  \\
    \bottomrule
  \end{tabular}
  
  \label{tab:acc_full_data}
\end{table}


\begin{table}[t!] 
  \centering
  \small
  \caption{Classification accuracy (ACC) on ImageNet dataset by competing approaches using different CNN backbones.}
  \begin{tabular}{l@{\;\,}c@{\qquad\;}l@{\;\,}c}
    \toprule
    \makecell{\bf CNN \\ \bf Model} & \textbf{ACC} & \makecell{\bf Transformer \\ \bf Model} & \textbf{ACC} \\
    \midrule
    ResNet-34 & 73.3\% & ViT-B/16 & 81.1\%  \\
    \hspace{5pt}\includegraphics[width=6pt]{images/subdirectory_arrow_right.png}\,\our{} & 64.1\% & \hspace{5pt}\includegraphics[width=6pt]
    {images/subdirectory_arrow_right.png}\,\our{} & 79.2\% \\
    \cdashlinelr{1-4}
    ResNet-50 & 76.1\%  & Swin-S & 83.4\%  \\
    \hspace{5pt}\includegraphics[width=6pt]{images/subdirectory_arrow_right.png}\,\our{} & 67.8\% & \hspace{5pt}\includegraphics[width=6pt]{images/subdirectory_arrow_right.png}\,\our{} & 81.4\% \\
    \cdashlinelr{1-4}
    DenseNet-121 & 74.4\%  & MaxVit & 83.4\% \\
    \hspace{5pt}\includegraphics[width=6pt]{images/subdirectory_arrow_right.png}\,\our{} & 66.6\% & \hspace{5pt}\includegraphics[width=6pt]{images/subdirectory_arrow_right.png}\,\our{} & 83.3\% \\
    \cdashlinelr{1-4}
    ConvNeXt-L & 84.1\%  & & \\
    \hspace{5pt}\includegraphics[width=6pt]{images/subdirectory_arrow_right.png}\,\our{} & 82.8\% &  &  \\
    \bottomrule
  \end{tabular}
  
  \label{tab:acc_imagenet}
\end{table}

\subsection{Classification Performance} 

To compare our approach, we selected several state-of-the-art, interpretable models based on the same CNN architectures as our approach. We categorized the models into a few groups based on their CNN architecture -- ResNet-34/50~\cite{he2016deep}, DenseNet-121~\cite{huang2017densely}, and ConvNeXt-Tiny~\cite{liu2022convnet} -- and the experiments we conducted on these datasets. Specifically, we performed two experiments: the first used cropped images for training and testing, and the second used full images.

In the first experiment, we utilized two key datasets: CUB-200-2011 and Stanford Cars, which are frequently employed in prototype model evaluations. We trained both the base models and our models on these cropped images, with the results shown in~\Cref{tab:acc_cropped_data}. In the second experiment, we used CUB-200-2011 and Stanford Dogs datasets, but this time with the full images. The results of this experiment are detailed in~\Cref{tab:acc_full_data}.

In both experiments, our approach consistently outperformed the compared models. Additionally, training competitive methods often proves to be more challenging, requiring substantial computational resources and time due to the complex training processes involved. In contrast, our method simplifies this by only training the last two parts of the model network, with the CNN backbone remaining unchanged, thereby reducing the training effort and complexity.

In the next experiment in this part, we utilized the full ImageNet dataset and evaluated both traditional CNN models -- such as ResNet-34/50, DenseNet-121, and ConvNeXt-Large -- and popular transformer models, including VisionTransformer (ViT-B/16)~\cite{dosovitskiy2020image}, SwinTransformer (Swin-S)~\cite{liu2022swin}, and MaxVit~\cite{tu2022maxvit}. Given that current prototype models did not perform well on the ImageNet dataset and thus lack evaluation results, \Cref{tab:acc_imagenet} presents a comparison between the classical models and our approach.

Typically, prototype models exhibit lower performance on more complex datasets compared to traditional models. However, as shown in~\Cref{tab:acc_imagenet}, our approach not only matches the performance of these classical models but also offers enhanced explainability.

\begin{figure*}[t]  
    \centering
    \includegraphics[width=0.98\linewidth]{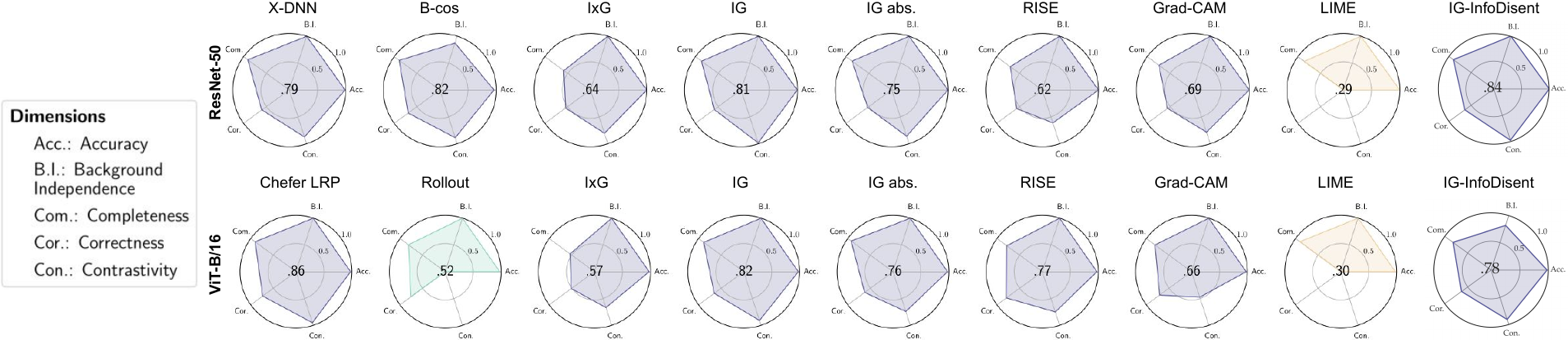}
    \caption{
    FunnyBirds evaluation results for various XAI methods, 
    Model-agnostic methods are assessed on ResNet-50 and ViT-B/16. Results are averaged over the entire test set, including the center score representing the mean of completeness (Com.), correctness (Cor.), and contrastivity (Con.) dimensions. Additionally, accuracy (Acc.) and background independence (B.I.) are reported. Our approach (at the end on the left) enhances model explainability, showing significant improvements for ResNet-50 and satisfactory results for the transformer model. 
    }
    \label{fig:FunnyBirds}
\end{figure*}

\section{Evaluation of interpretability}

We showcase the performance of \our{} on explainability metrics using FunnyBirds dataset~\cite{hesse2023funnybirds} which is a dedicated benchmark for XAI methods.

In addition to the computational validation of \our{}, we conducted two user studies. The first study followed the design of the HIVE benchmark~\cite{kim2022hive}, aiming to assess the impact of explanations on user overconfidence in the model's predictions. The second study evaluated the disambiguation of explanations derived from prototypical parts. For this, we adopted a study design from prior works on prototypical parts~\cite{ma2023looks,pach2024lucidppn}.

\paragraph{FunnyBirds results}
In the last experiment, we utilized the FunnyBirds~\cite{hesse2023funnybirds} dataset to evaluate our approach. The FunnyBirds dataset, along with our novel automatic evaluation protocols, supports semantically meaningful image interventions, such as removing individual object parts. This enables a more nuanced analysis of explanations at the part level, which aligns more closely with human understanding compared to pixel-level evaluations. Results spanning a range of XAI methods and two different types of XAI model architectures are detailed in~\Cref{fig:FunnyBirds}. \our{} enhances the explainability of models based on classic CNN architectures like ResNet-50 and ranks among the top XAI methods for transformers.

\begin{table}[t]\small
    \centering
    \caption{Results on user confidence in model predictions show that users, when answering questions based on explanations from \our{} model, are generally confident in the correctness of the model's predictions for both datasets ($>0.6$ for ImageNet and $>0.8$ for CUB). However, users face challenges in identifying incorrect predictions made by the model based on the explanations. This observation aligns with findings for other XAI techniques. Additionally, the results for \our{} are statistically significant. We bold statistically significant values (p-value $<0.05$). Note that results for ProtoPNet and GradCAM are referenced from the original HIVE study~\cite{kim2022hive}, as indicated by the asterisk symbol.}
    \renewcommand{\arraystretch}{1.2} 
    \begin{tabular}{@{}cccc@{}}
    \toprule
        Method & Prediction & ImageNet & CUB-200-2011 \\
    \midrule
        \multirow{2}{*}{\our{}} & Correct & $\mathbf{0.602 \pm 0.090}$ & $\mathbf{0.807 \pm 0.133}$ \\
         & Incorrect & $\mathbf{0.553 \pm 0.099}$ & $\mathbf{0.427 \pm 0.117}$  \\
    \midrule
        \multirow{2}{*}{ProtoPNet$^*$} & Correct & NA & $0.732 \pm 0.249$  \\
        & Incorrect & NA & $\mathbf{0.464 \pm 0.359}$ \\
        \multirow{2}{*}{GradCAM$^*$} & Correct & $0.708 \pm 0.266$ & $0.724 \pm 0.215$\\
        & Incorrect & $\mathbf{0.448 \pm 0.316}$ & $0.328 \pm 0.243$ \\
    \bottomrule
    \end{tabular}
    \label{tab:hive}
\end{table}

\subsection{User study results}

We performed two user studies. Each of them involved 60 participants per dataset, with a balanced gender representation. Participants were aged between 18 and 60, with an average age of 35 years. The studies were conducted on the Clickworker platform, using two datasets: CUB-200-2011 and ImageNet. Each participant answered 20 questions, with images randomly selected from the testing dataset for each question. Example questions are provided in the Supplementary Materials.

\paragraph{Confidence in Model Predictions} In the first study, which assessed user overconfidence, participants were presented with an image alongside the model’s explanation. They were then asked, “What do you think about the model’s prediction?” and were instructed to indicate their confidence in whether the model was correct or incorrect.

The results of this user study are shown in~\Cref{tab:hive}. They demonstrate that users evaluating the model’s predictions based on explanations from \our{} are statistically significantly less overconfident than random guessing (as indicated by the p-values). Additionally, users perform comparably to ProtoPNet on CUB, while \our{} also extends the concept of prototypical parts to ImageNet.

\begin{table}[t]\small
    \centering
    \caption{User study results show that users, based on explanations from \our{}, perform statistically significantly better in understanding the model’s decisions for both ImageNet and CUB datasets than a random guessing. Note that \our{} is the only prototypical partslike method working on ImageNet.  Additionally, they achieve comparable performance to methods focused on disambiguating prototypical parts, such as ProtoConcepts and LucidPPN. Note that results for ProtoPNet and ProtoConcepts are referenced from~\cite{ma2023looks}, while results for PIPNet and LucidPPN are cited from~\cite{pach2024lucidppn}, as indicated by the asterisk symbol. The p-value column indicates the p-value of a test against random.}
    \renewcommand{\arraystretch}{1.2} 
    \begin{tabular}{@{}l@{\quad}cl@{\quad}cl@{\quad}c@{}}
    \toprule
        \multicolumn{1}{c}{Method} & Dataset & User Acc. & p-value \\
    \midrule
        \multirow{2}{*}{\our{}} & ImageNet & $\mathbf{0.593 \pm 0.149}$ & $\mathbf{8\cdot10^{-6}}$ \\
        & CUB & $0.647 \pm 0.131$ & $10^{-14}$ \\
    \midrule
        ProtoPNet$^*$ & \multirow{4}{*}{CUB} & $0.515 \pm 0.052$ & $0.288$ \\
        ProtoConcepts$^*$ & & $0.621 \pm 0.054$ & $3\cdot10^{-5}$ \\
        PIPNet$^*$ & & $0.600 \pm 0.181$ & $0.002$ \\
        LucidPPN$^*$ & & $0.679 \pm 0.169$ & $2\cdot10^{-6}$ \\
    \bottomrule
    \end{tabular}
    \label{tab:ambiguity}
\end{table}

\paragraph{Disambiguity of prototypical parts.} In the second study, which assessed the disambiguation of prototypical parts, participants were shown an image classified by the model alongside two explanations corresponding to the two most activated classes. Their task was to determine, based on the explanations, which decision the model had made.

The results of this user study, presented in ~\Cref{tab:ambiguity}, demonstrate that users who saw explanations from \our{} performed statistically significantly better than random guessing. Additionally, when compared to other methods focused on disambiguating prototypical parts, such as ProtoConcepts and LucidPPN, users based on the \our{} explanations achieves comparable performance.

\section{Conclusions}




In this work, we introduce \our{}, an innovative model that combines the strengths of both post-hoc and inherently interpretable methods. \our{} provides the flexibility to be applied to any backbone, while offering both local (per image) and global (per class) explanations in the form of atomic concepts, addressing key limitations of existing approaches. Additionally, as demonstrated by user studies, \our{} performs comparably to state-of-the-art methods in disambiguating prototypical parts and managing user overconfidence. Notably, \our{} is the first attempt to generalize the prototypical parts-based methodology to big scale datasets such as the whole ImageNet. In future work, we plan to explore pruning techniques to optimize the size of concepts used in explanations.

\paragraph{Limitations} To achieve interpretability, \our{} requires training its head on the entire dataset. Furthermore, the model’s decisions are not reduced to a fixed number of prototypes, unlike prototypical parts models.

\paragraph{Impact} \our{} advances the field of Explainable AI (XAI) by introducing a novel method that generalizes prototypical parts-based explanations to ImageNet-like datasets, while maintaining the post-hoc flexibility of application. \our{} holds potential for further exploration in downstream applications, such as medical diagnosis.

\paragraph{Acknowledgments} This research was partially funded by the National Science Centre, Poland, grants no.  2020/39/D/ST6/01332 (work by \L{}ukasz Struski), and 2023/49/B/ST6/01137 (work by Dawid Rymarczyk  and Jacek Tabor). Some experiments were performed on servers purchased with funds from the flagship project entitled ``Artificial Intelligence Computing Center Core Facility'' from the DigiWorld Priority Research Area within the Excellence Initiative -- Research University program at Jagiellonian University in Kraków.

\bibliographystyle{plain}
\bibliography{egbib}


\newpage
\appendix
\onecolumn

\begin{minipage}[t]{\textwidth}
\begin{center}
\LARGE\bf
Supplementary Material
\end{center}
\vspace{2em}
\end{minipage}

\section{Ablation study}
\our{} organizes information into channels with sparse representations, which can be later utilized in the model's prediction process. In the following experiment, we investigate how the number of channels affects the model's predictions. Specifically, we assess how many channels are required to account for at least 95\% of the information used in the model’s predictions. Formally, if 
\(
\text{logits} = \sum\nolimits_{i=0}^N a_{ki} v_i + b_k,
\)
where \(N\) is a number of all channels, \(k\) represents the image class, and \(a_{ki}, v_i, b_k \in\R\), then for each image from class \(k\), we determine the smallest number \(n\) channels such that 
\(
\sum\nolimits_{i \in I_k} |a_{ki} v_i| / \sum\nolimits_{i=0}^N |a_{ki} v_i| \geq 0.95,
\)
where \(I_k\) is the set of indexes of the \(n\leq N\) largest values of \(|a_{ki} v_i|\). This analysis allows us to identify the most critical channels contributing to the model's decisions, providing deeper insights into the model's interpretability and efficiency.

\begin{figure}[htb]
  \centering
  \small
   \includegraphics[width=.8\textwidth]{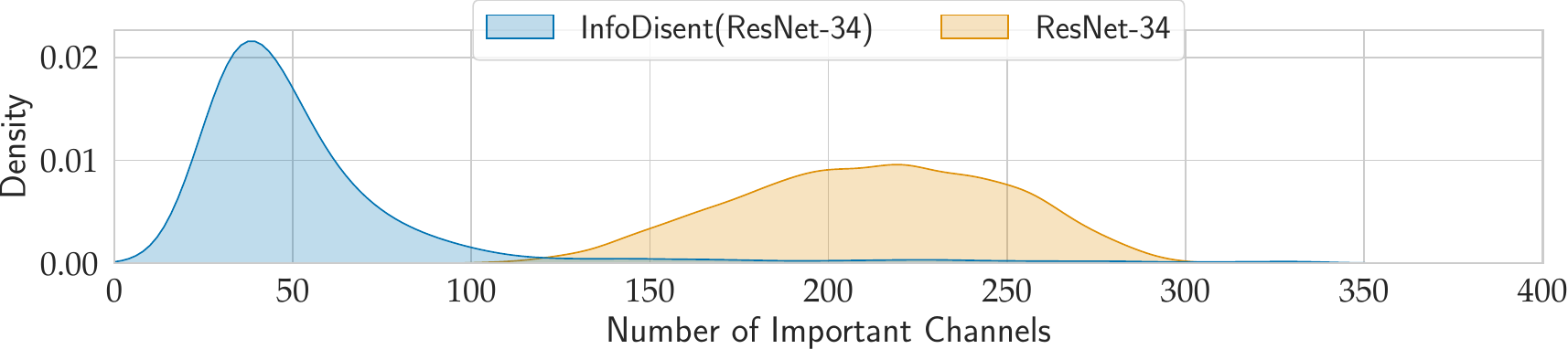} 
  \caption{Density estimate of the number of significant channels for each class and image in the CUB-200-2011 test set using the ResNet-34 network. \our{} uses significantly fewer channels and has less variance.
  }
  \label{fig:decision_behind_class}
\end{figure}

\Cref{fig:decision_behind_class} shows the density estimate of the number of significant channels for each class and image in the CUB-200-2011 test set. In this experiment, we used the ResNet-34 network that utilizes a significantly larger number of channels in its predictions compared to \our{} model. 
By reducing the number of significant channels while preserving classification performance, our model demonstrates efficient resource utilization and improved interpretability, as illustrated below. This efficiency shows that our model is more effective at isolating the critical features necessary for accurate predictions, thereby validating our approach.
This also validates the disentangling role of the orthogonal matrix $U$.

\section{Architectural Adjustments for Training}

In this section, we will provide a detailed explanation of the training process for the \our{} network. Recall that the first component of the network is the CNN or Transformer Backbone, a pre-trained classical model that is frozen up to the final feature mapping layer. During \our{} training, this backbone, along with its weights, remains unchanged, preserving the learned feature representations.

\begin{figure}[t]
\centering
\includegraphics[width=0.9\linewidth]{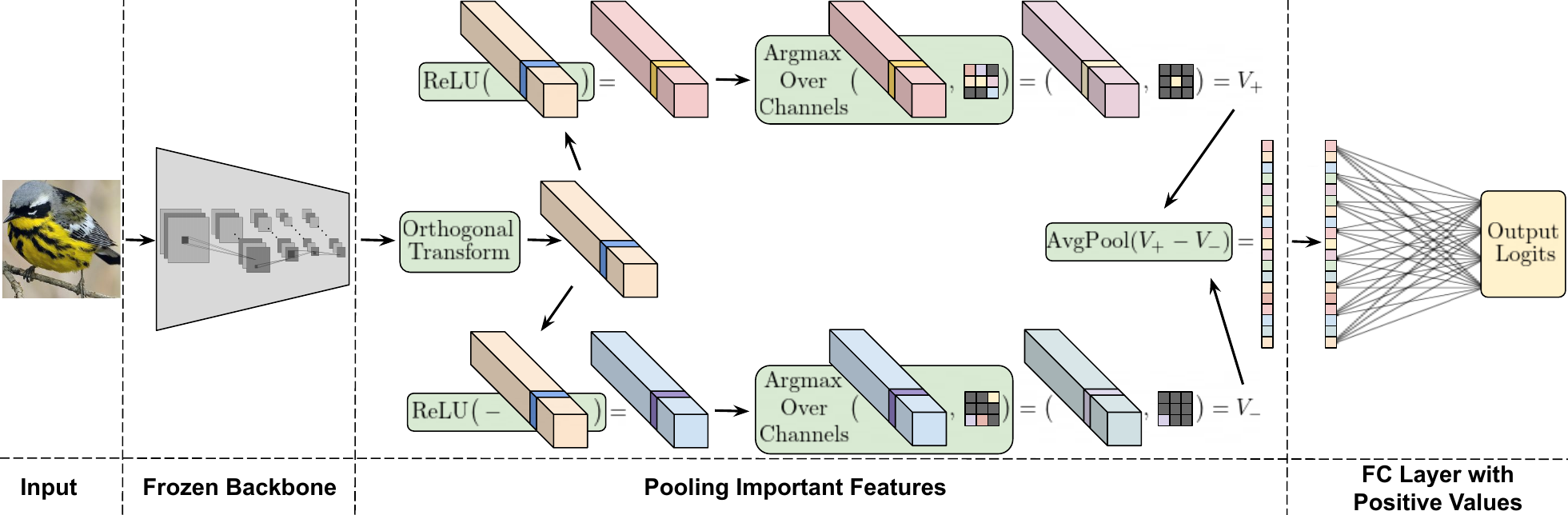}
\captionof{figure}{The architecture used for training of our proposed image classification interpretation model. \our{} is composed of three main components: a pre-trained backbone, a pooling layer for extracting important features, and a fully connected layer. The backbone is a pre-trained CNN or transformer with frozen weights, meaning it is not further trained. In the initial pooling layer, the model extracts representations from the last convolutional layer of the backbone and identifies key features within each channel, targeting both positive and negative activations through the application of the $\argmax$ operation. However, during training, we replace the $\argmax$ operation with the Gumbel-Softmax trick, which achieves a similar outcome in a differentiable manner. In the next step, these positive and negative features are pooled at the channel level to create a dense vector, where the vector’s dimensions correspond to the number of channels. Finally, this dense vector is passed through a fully connected linear layer with positive weights in the network's final component.}
\label{fig:architecture}
\end{figure}

The second component of the network involves combining essential features from the final feature mapping produced by the Frozen Backbone. This combination mechanism encourages the network to leverage information bottlenecks, thereby constructing maximally informative and independent channels. These channels, known as prototype channels, are designed to be easily interpretable. The final component is a fully connected layer characterized by positive weight values.

To maximize the extraction of information from these channels, we introduce an information bottleneck within our model architecture. This is achieved by applying the $\argmax$ operation to individual channels, see~\Cref{fig:architecture}. While the $\argmax$ function can be used to extract a sparse representation from feature maps, our goal is to enable the model to learn to select the most important values. To achieve this, we require a differentiable $\argmax$ function. The ideal solution for this is the Gumbel-Softmax estimator~\cite{jang2016categorical}. Given $x = (x_1, \ldots, x_D) \in\R^D$ and $\tau\in(0, \infty)$,
\[
\mathrm{Gumbel\textnormal{-}Softmax}(x, \tau) = (y_1, \ldots, y_D) \in \mathbb{R}^D,
\]
where
\[
y_i = \frac{\exp\left((x_i + \eta_i) / \tau\right)}{\sum_{d=1}^D \exp\left((x_d + \eta_d) / \tau\right)},
\]
and $\eta_d$ for $d\in\{1, \ldots, D\}$ are samples taken from the standard Gumbel distribution.

The Gumbel-Softmax distribution serves as an interpolation between continuous categorical densities and discrete one-hot encoded categorical distributions, with the discrete form being approached as the temperature $\tau$ decreases within the range of $[0.1, 0.5]$. In our experiments, we initialized $\tau$ at 1 and progressively reduced it to 0.2. Finally, at the end of the training, we applied a hard softmax.

Following the extraction of key features using the sparse operation -- specifically, the $\argmax$ operation via the Gumbel-Softmax trick -- we preserve the original structure of the network's output, maintaining the classical form of the convolutional network's output, as shown in~\Cref{fig:architecture}. During the subsequent aggregation of positive and negative features, we utilize an average pooling operation to consolidate the information. This approach ensures that the pooled features capture a balanced representation of the activations, contributing to a robust final output.

\section{Details of the experiments performed}

\paragraph{Datasets} In our experiments, we leveraged several diverse datasets to evaluate performance. The first dataset is the Caltech-UCSD Birds-200-2011 (CUB-200-2011)\cite{wah2011caltech}, which contains 11,788 images meticulously labeled across 200 bird species, divided into 200 subcategories. Of these, 5,994 images are allocated for training, while 5,794 are reserved for testing. The second dataset, known as Stanford Cars\cite{krause20133d}, is designed to classify various car models. It includes 16,185 images, each capturing a rear view of different car types across 196 classes, with an almost even distribution between training (8,144 images) and testing (8,041 images) subsets. Each class details the car's make, model, and year. The third dataset, Stanford Dogs~\cite{khosla2011novel}, features a collection of 20,580 images representing 120 dog breeds from around the globe. This dataset, sourced and annotated through ImageNet, is intended for fine-grained image classification, with 12,000 images for training and the remainder for testing.

Additionally, we incorporated the FunnyBirds~\cite{hesse2023funnybirds} dataset, consisting of 50,500 images representing 50 synthetic bird species, with 50,000 images for training and 500 for testing. This dataset was designed with a focus on "concepts," or mental representations crucial for categorization, and is particularly relevant for explainable AI (XAI). The concepts are linked to specific bird anatomy parts, such as the beak, wings, feet, eyes, and tail, ensuring they are both granular and intuitive for practical use in XAI.

Finally, we utilized ImageNet~\cite{russakovsky2015imagenet}, a highly recognized dataset in computer vision, often employed for pretraining deep learning models. ImageNet encompasses 1,281,167 training images, 50,000 validation images, and 100,000 test images, spanning 1,000 object classes.

\paragraph{Training Details} We train the architectures using stochastic gradient descent (SGD) with standard categorical cross-entropy loss. The momentum, damping, and weight decay are set to 0.9, 0.9, and 0.001, respectively. For the baseline networks, the initial learning rates are 0.1, 0.05, and 0.01, which are reduced by a factor of 0.1 when the validation loss converges. In our approach, we train only the last two segments of the network, thus we use lower learning rates of 0.001 and 0.0001, utilizing the 'ReduceLROnPlateau'~\cite{al2022scheduling} mechanism that reduces the learning rate when the cost function stops improving. All numerical experiments were conducted using NVIDIA RTX 4090 and NVIDIA A100 40 GB graphics cards.

For cropped images, we follow previous studies~\cite{chen2019looks} by applying on-the-fly data augmentations (e.g., random rotation, skew, shear, and left-right flip) on the cropped CUB and cropped Cars datasets using the provided bounding boxes.
We also validate our method on the full (uncropped) CUB and Dogs datasets, employing the same online data augmentation techniques (e.g., random affine transformation and left-right flip). For the FunnyBirds dataset, we adhered to the detailed instructions provided in the framework's documentation, which can be found at \url{https://github.com/visinf/funnybirds}. For training various CNN and transformer models on the ImageNet dataset, we utilized the augmentation techniques described at \url{https://github.com/pytorch/vision/tree/main/references/classification}.

\begin{figure}[t!] 
  \centering
  \includegraphics[width=.75\linewidth]{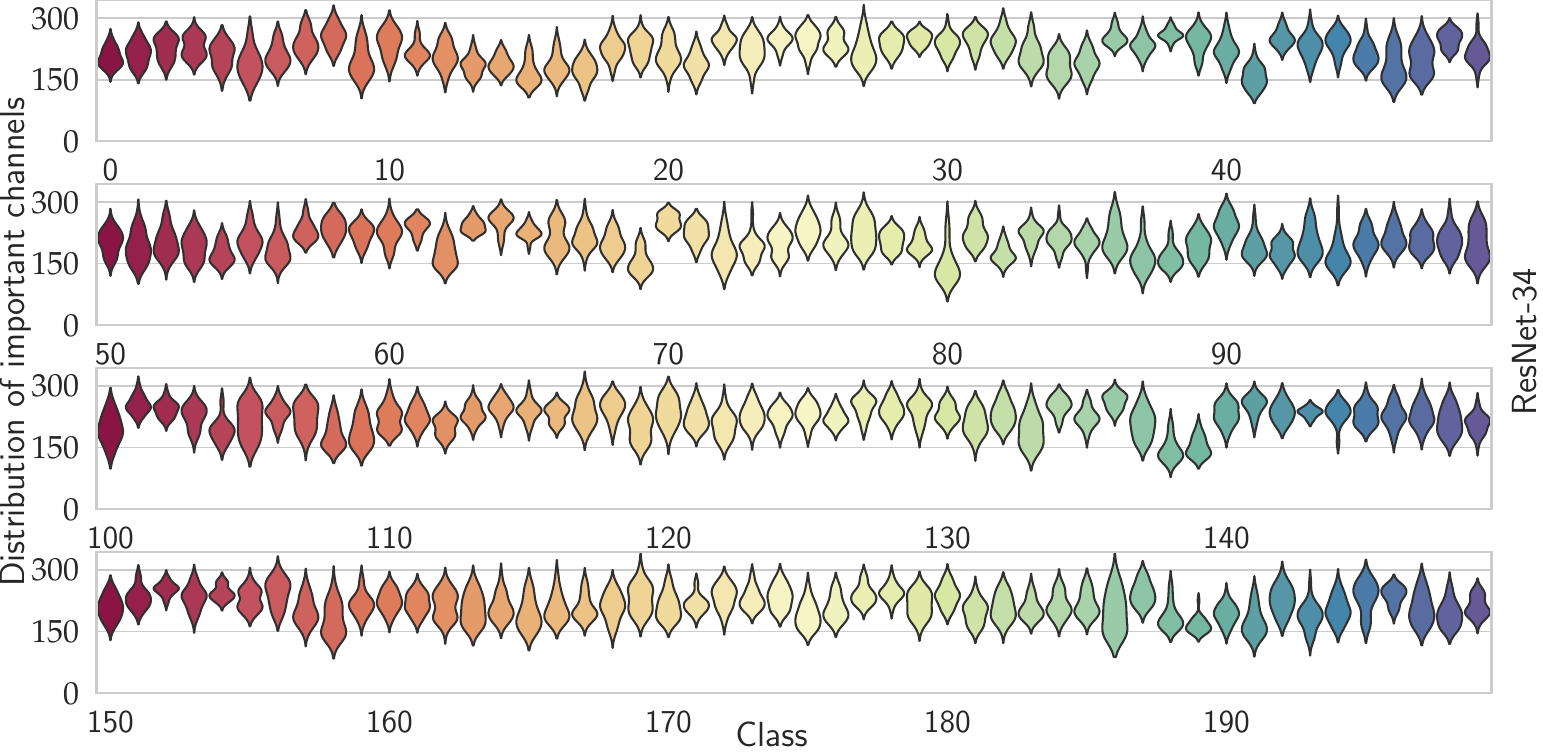} \\
   \includegraphics[width=.75\linewidth]{images/distribution_important_channels_own_resnet34_cub_200.pdf} 
  \includegraphics[width=.75\linewidth]{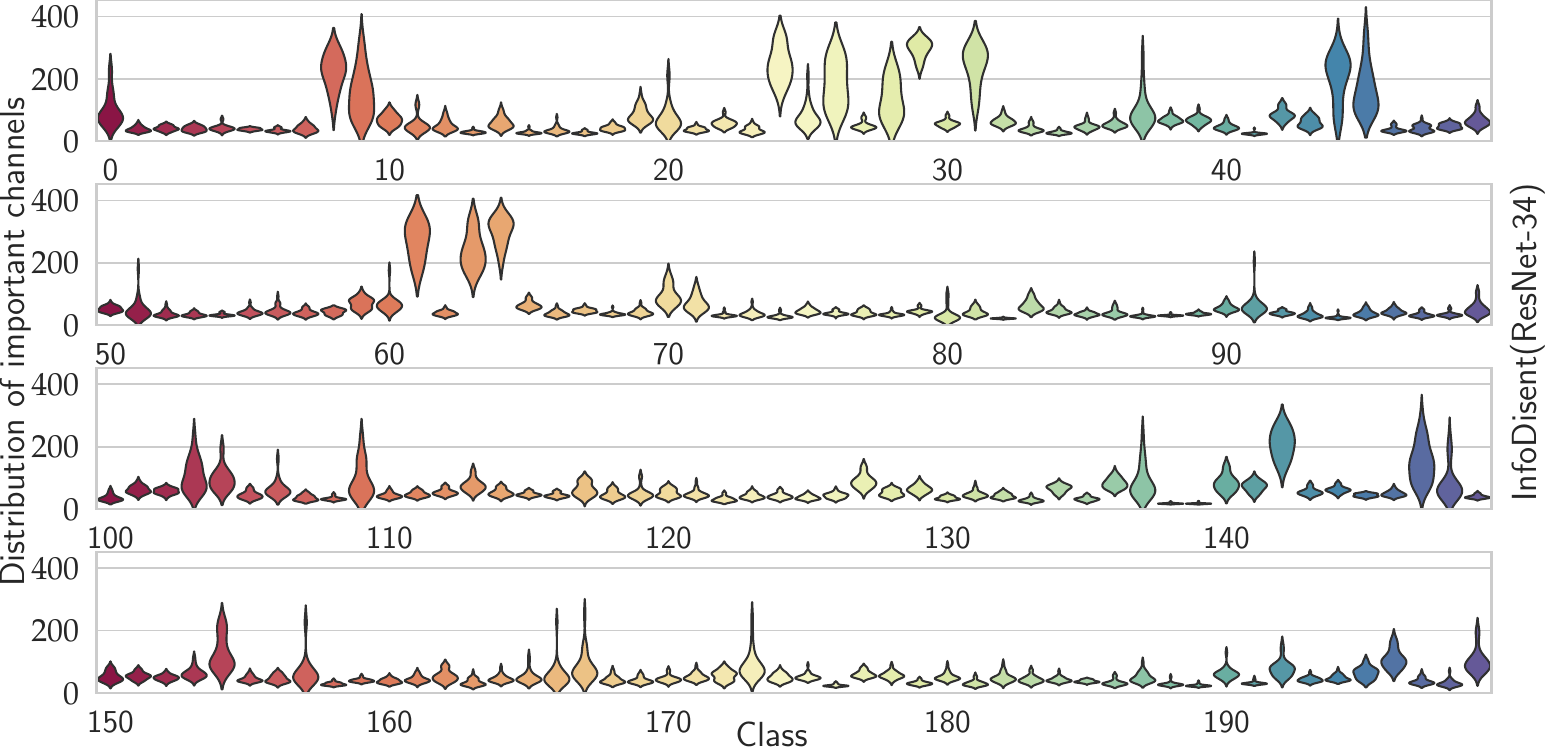}
  \caption{Density estimate of the number of significant channels for each class and image in the CUB-200-2011 test set using the ResNet-34 network. \our{} uses significantly fewer channels, particularly evident in the middle image, which shows the density estimation of the number of significant channels for all classes.
  }
  \label{fig:decision_behind_class_r34_birds}
\end{figure}

\begin{figure}[thb]
  \centering
  \includegraphics[width=.8\linewidth]{images/distribution_important_channels_own_resnet34_cub_200.pdf} \\
  \includegraphics[width=.8\linewidth]{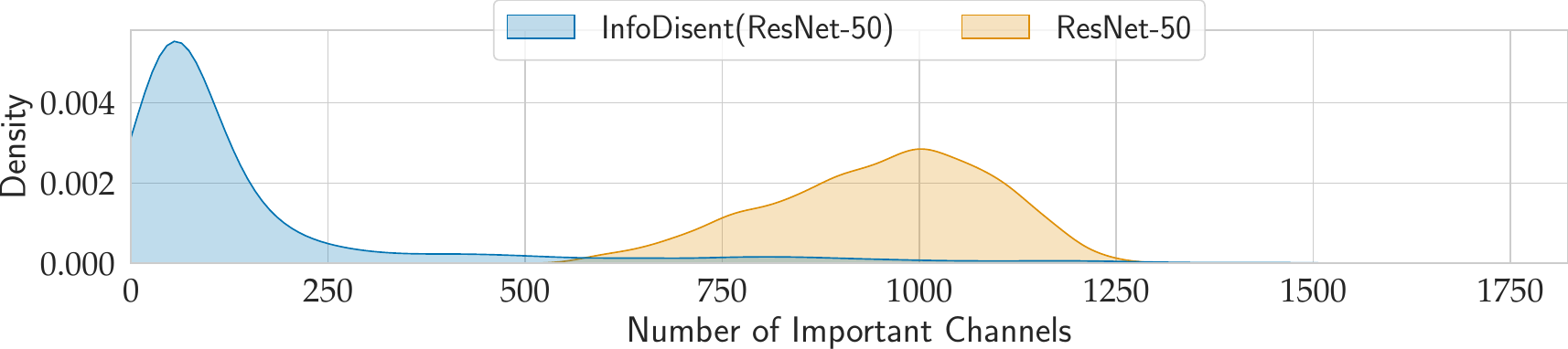} \\
  \includegraphics[width=.8\linewidth]{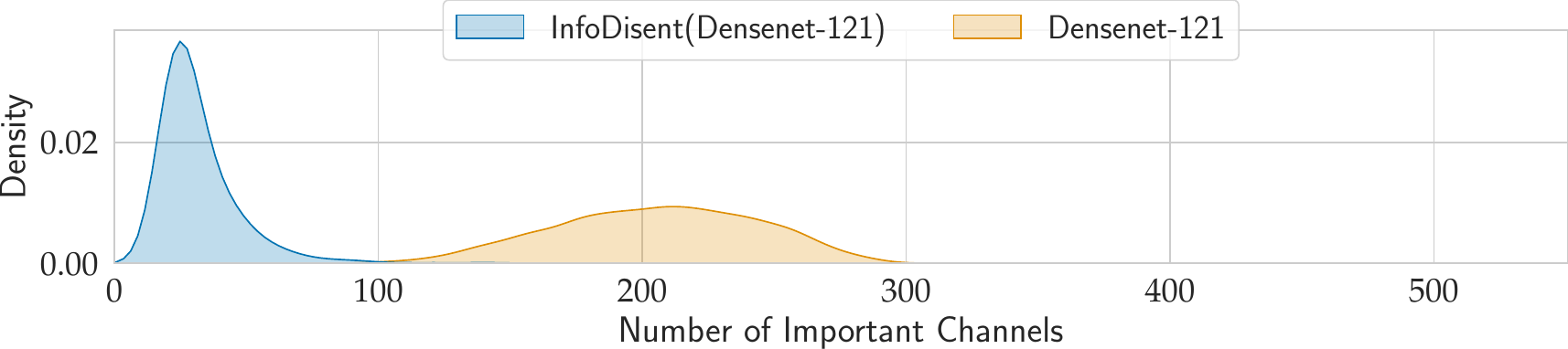} \\
  \includegraphics[width=.8\linewidth]{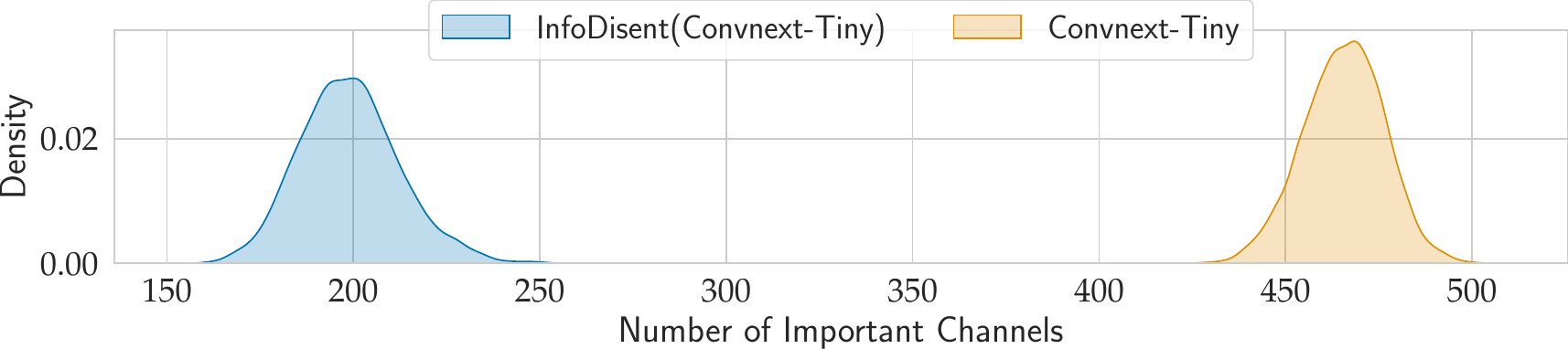}
  \caption{Density estimate of the number of significant channels for each class and image in the CUB-200-2011 test set using various networks.}
  \label{fig:decision_behind_class_birds}
\end{figure}

\begin{figure}[thb]
  \centering
  \includegraphics[width=.8\linewidth]{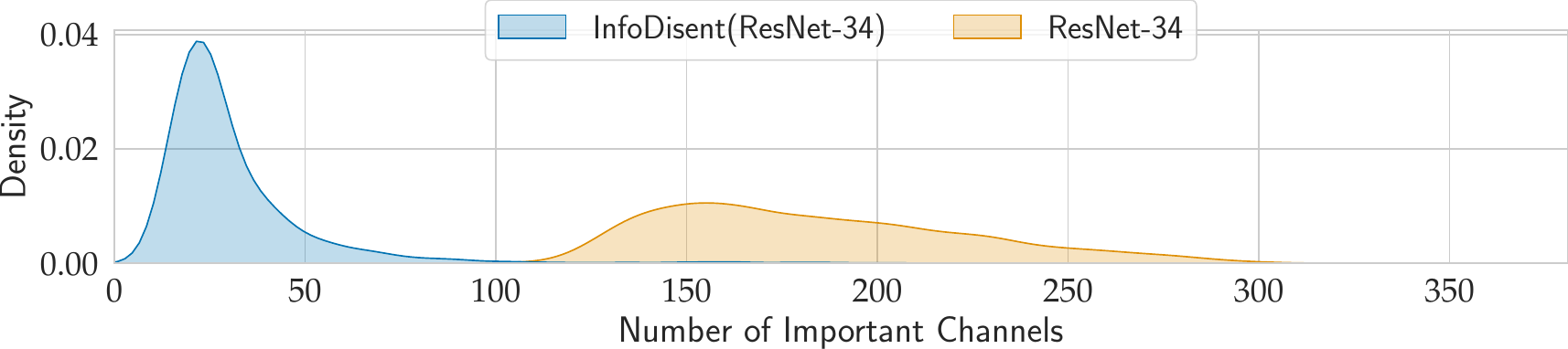} \\
  \includegraphics[width=.8\linewidth]{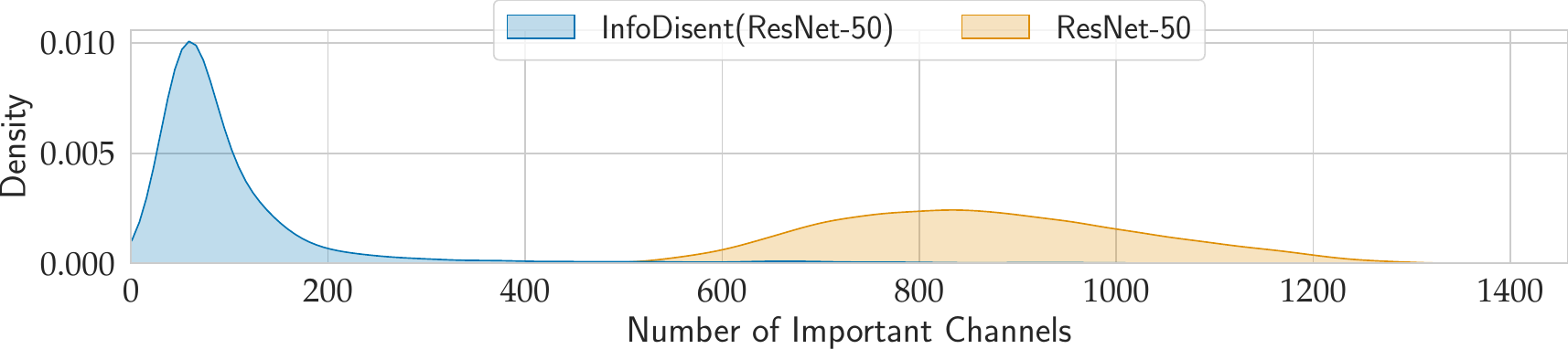} \\
  \includegraphics[width=.8\linewidth]{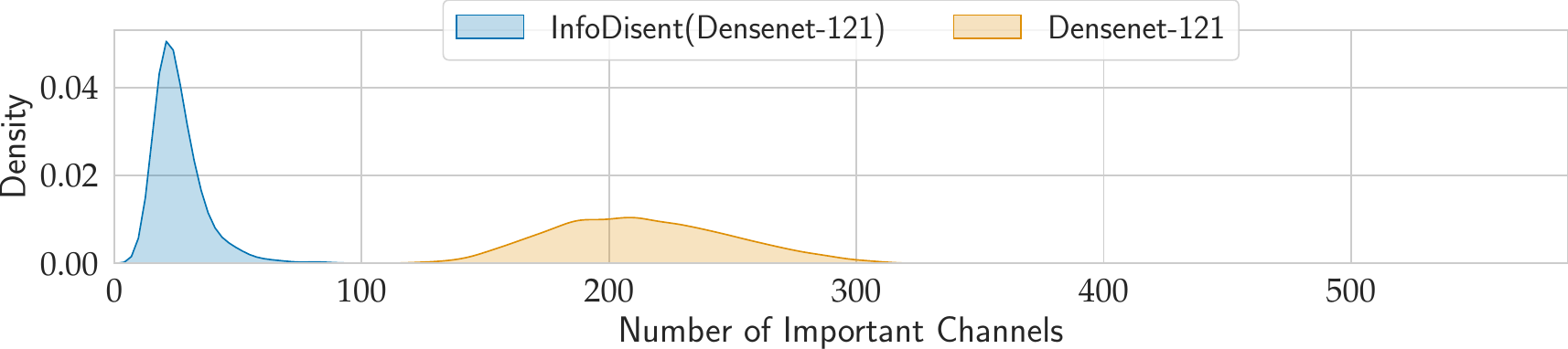} \\
  \includegraphics[width=.8\linewidth]{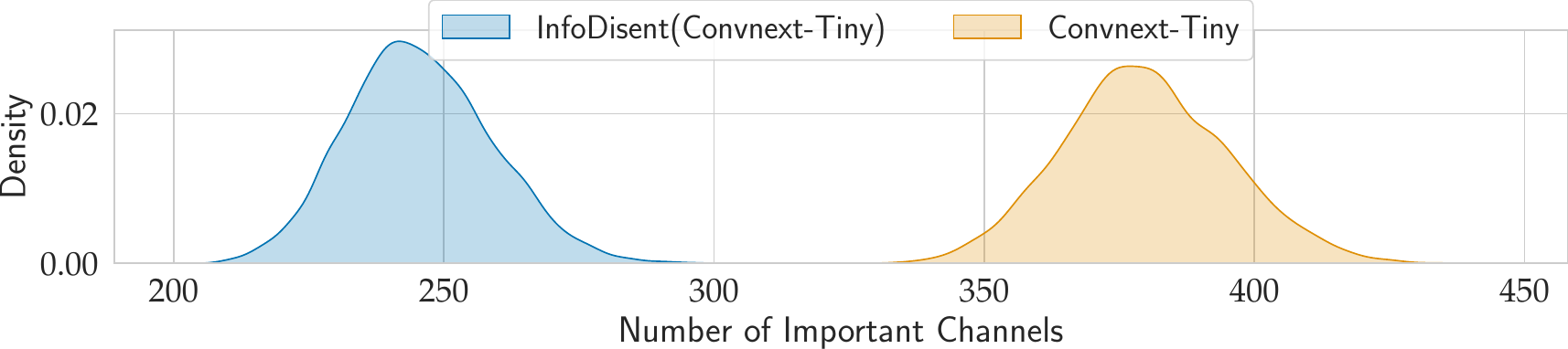}
  \caption{Density estimate of the number of significant channels for each class and image in the Stanford Cars test set using various networks.}
  \label{fig:decision_behind_class_cars}
\end{figure}

\section{Additional results}
\label{sec:add_results}

In this section, we provide additional results to supplement and expand upon the findings presented in the main part of the paper. We delve deeper into the behavior of our model across various experiments and datasets, offering a more comprehensive analysis.

\begin{figure}[t!]  
  \centering
  \includegraphics[width=.8\linewidth]{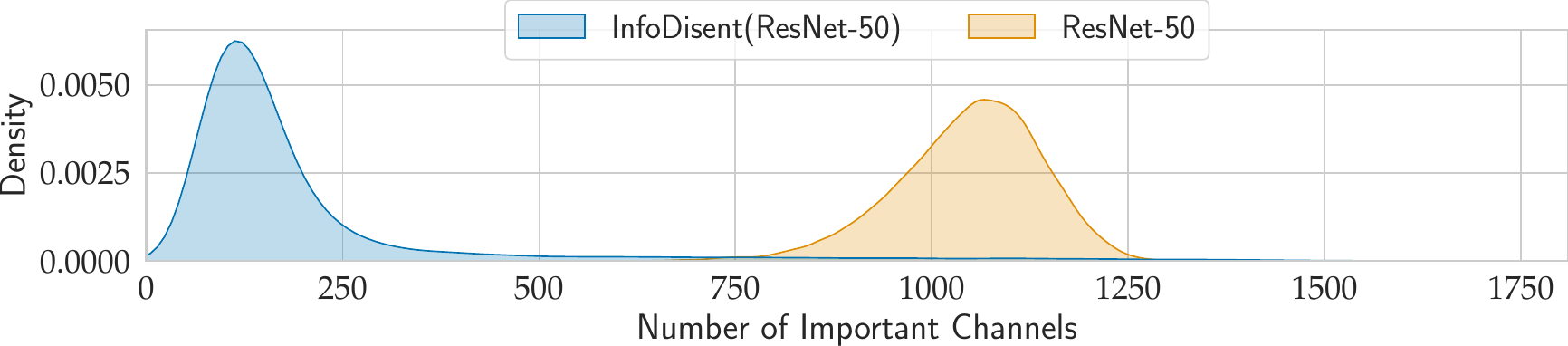} \\
  \includegraphics[width=.8\linewidth]{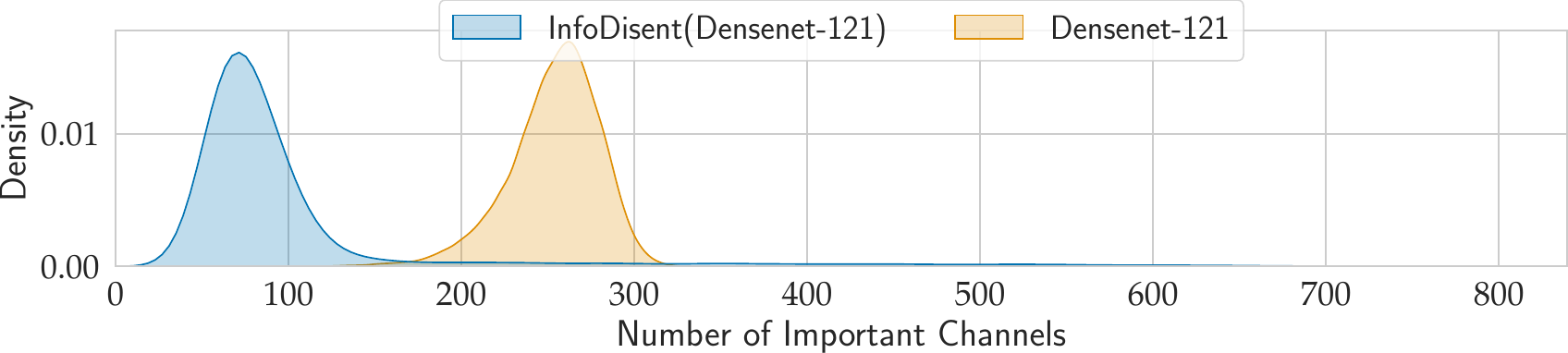} \\
  \includegraphics[width=.8\linewidth]{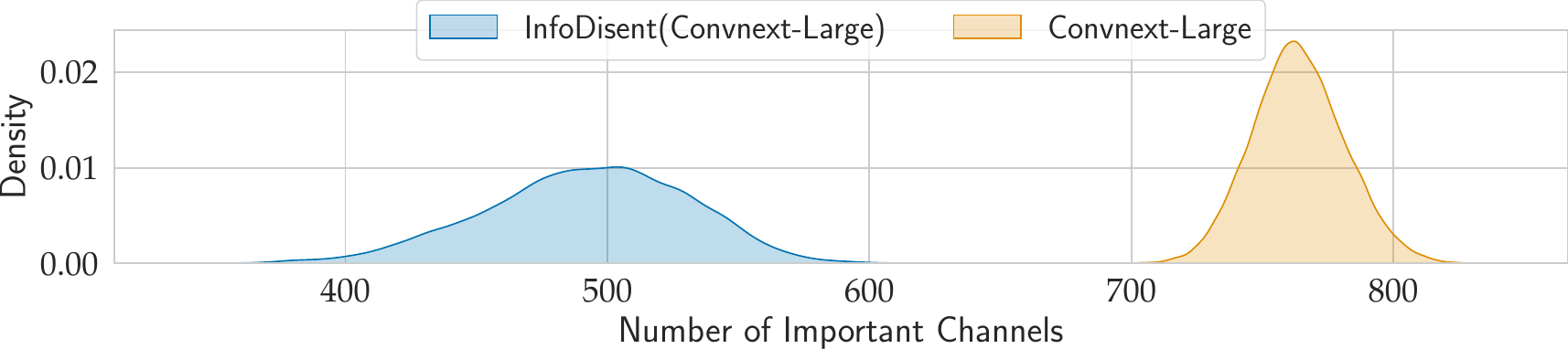} \\
  \includegraphics[width=.8\linewidth]{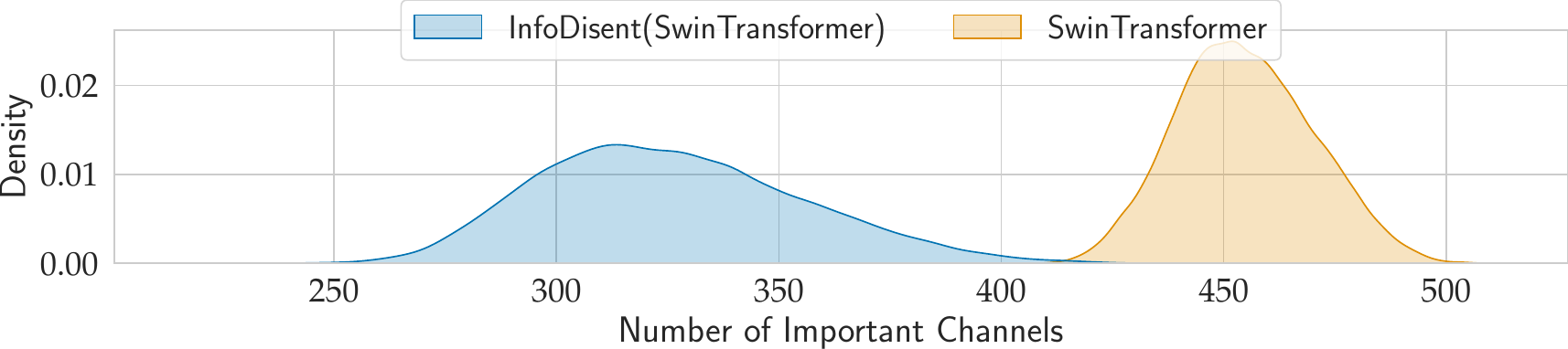} \\
  \includegraphics[width=0.8\linewidth]{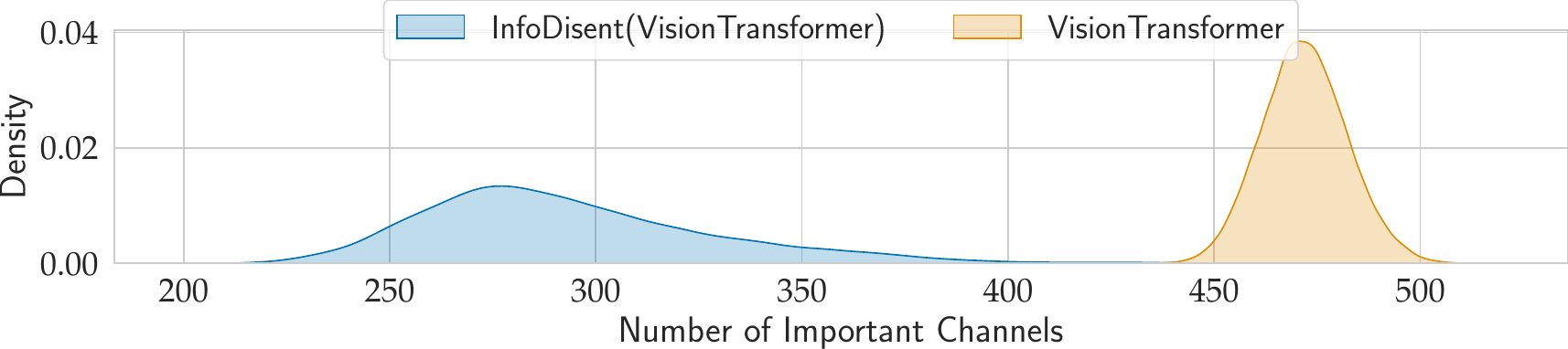}
  \caption{Density estimate of the number of significant channels for each class and image in the ImageNet test set using various networks.}
  \label{fig:decision_behind_class_imagenet}
\end{figure}

\paragraph{Ablation study}
In this part, we present additional results from a series of analyses investigating the significance of the number of channels on model predictions across various datasets and models. The results of these analyses are illustrated in~\Cref{fig:decision_behind_class_r34_birds,fig:decision_behind_class_birds,fig:decision_behind_class_cars,fig:decision_behind_class_imagenet}. Recall, that \our{} model organizes information into channels with sparse representations, which are later utilized in the model's prediction process. We specifically examine how the number of channels influences the model's predictions by determining the minimum number of channels required to account for at least 95\% of the information used in the model’s predictions.

Formally, if 
\[
\text{logits} = \sum_{i=0}^N a_{ki} v_i + b_k,
\]
where \(N\) is the total number of channels, \(k\) represents the image class, and \(a_{ki}, v_i, b_k \in \mathbb{R}\), then for each image from class \(k\), we identify the smallest number \(n\) of channels such that 
\[
\frac{\sum_{i \in I_k} |a_{ki} v_i|}{\sum_{i=0}^N |a_{ki} v_i|} \geq 0.95,
\]
where \(I_k\) is the set of indexes of the \(n\leq N\) largest values of \(|a_{ki} v_i|\). 

This analysis highlights the most critical channels contributing to the model's decisions, providing deeper insights into the model's interpretability and efficiency. Note that the \our{} approach consistently utilizes significantly fewer channels, a trend observed across all models and datasets analyzed.

\begin{figure*}[t!]  
  \centering
  \includegraphics[width=.9\linewidth]{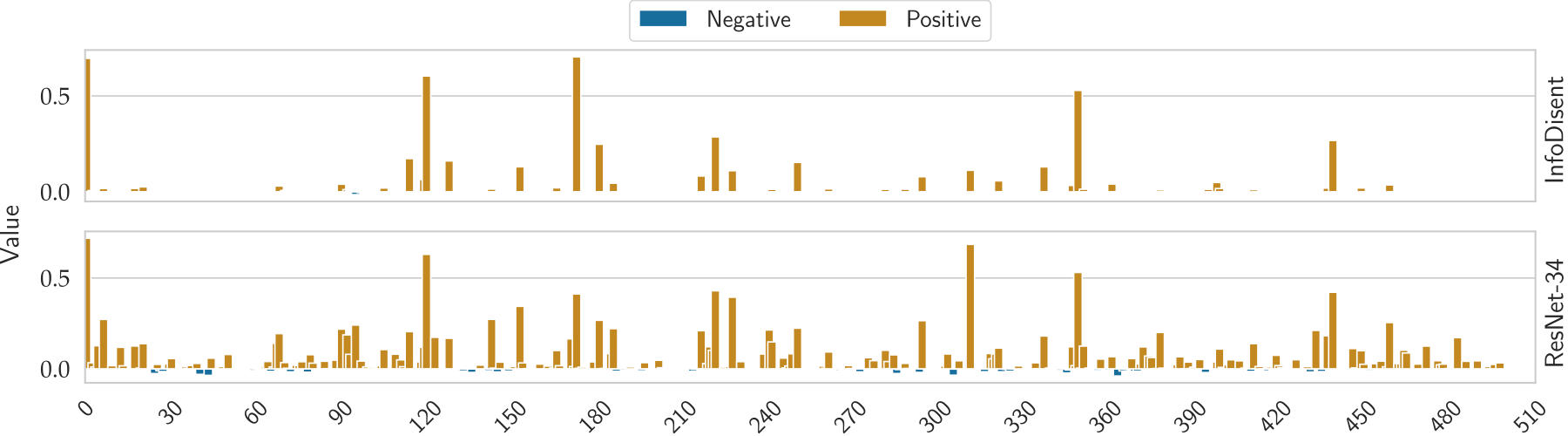} \\
  \includegraphics[width=.9\linewidth]{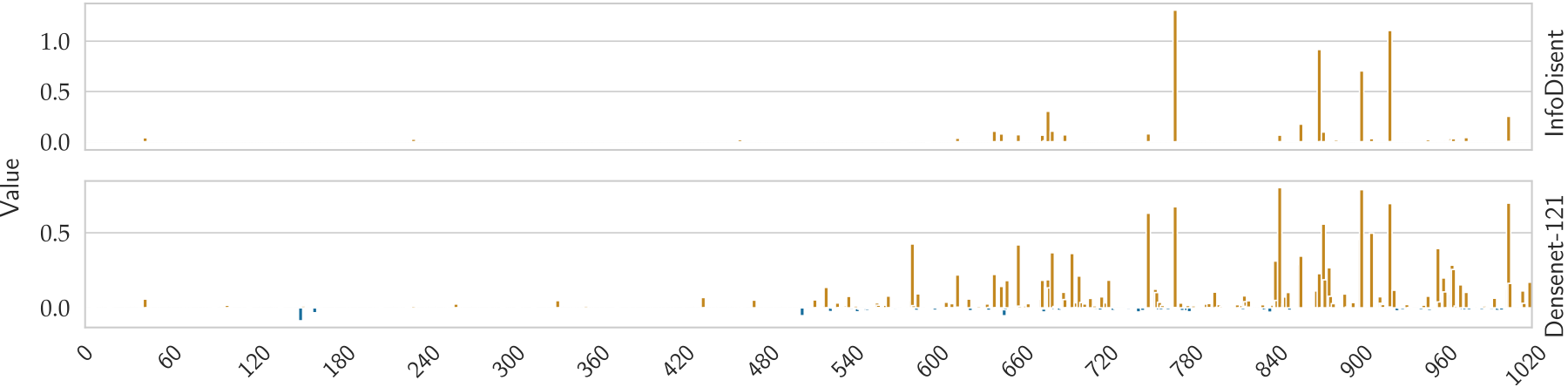} \\
  \includegraphics[width=.9\linewidth]{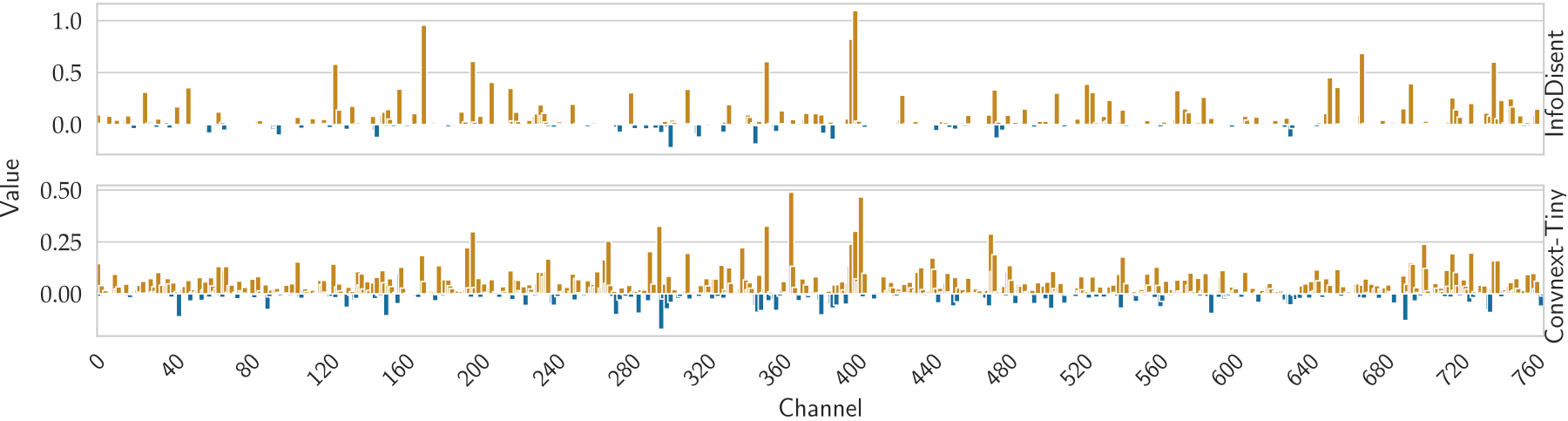}
  \caption{Channel activations before the final linear layer for a randomly selected image processed by different models trained on the CUB-200-2011 dataset are shown. The results are displayed in three groups, each containing two graphs. Model names are listed on the right side of the graphs. Unlike the baseline models, our network activates significantly fewer channels while still maintaining strong performance.}
  \label{fig:image_activation_channels_birds}
\end{figure*}

\begin{figure*}[t!] 
  \centering
  \includegraphics[width=.9\linewidth]{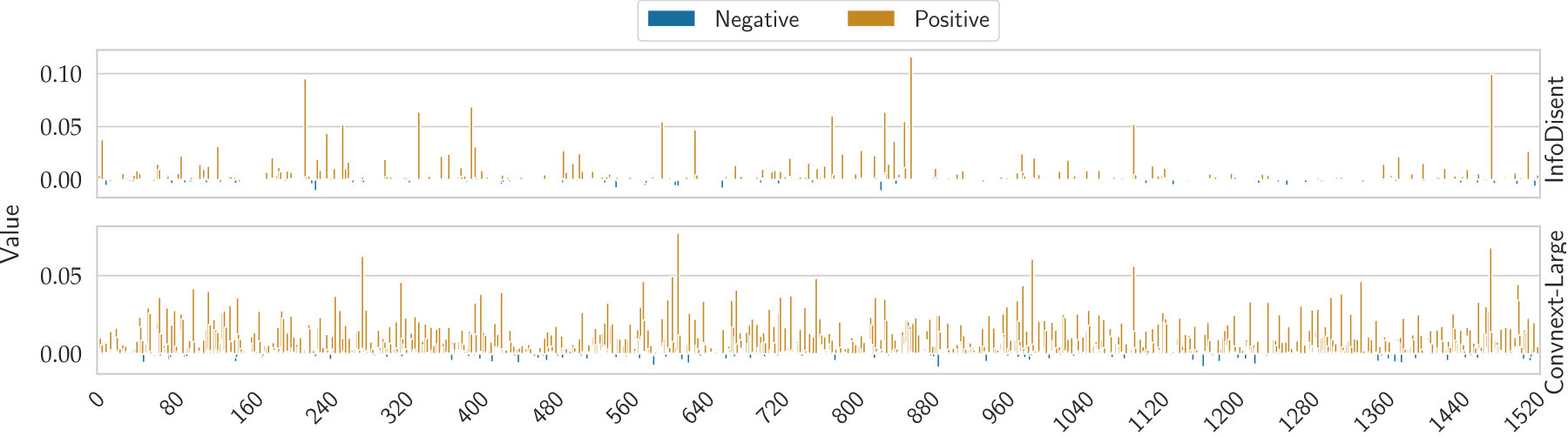} \\
  \includegraphics[width=.9\linewidth]{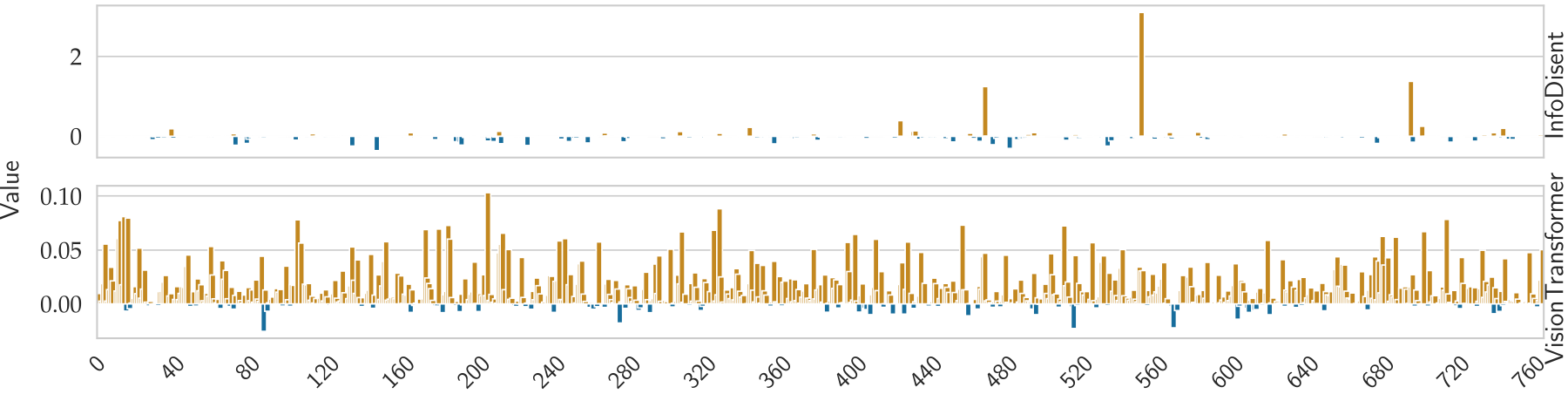} \\
  \includegraphics[width=.9\linewidth]{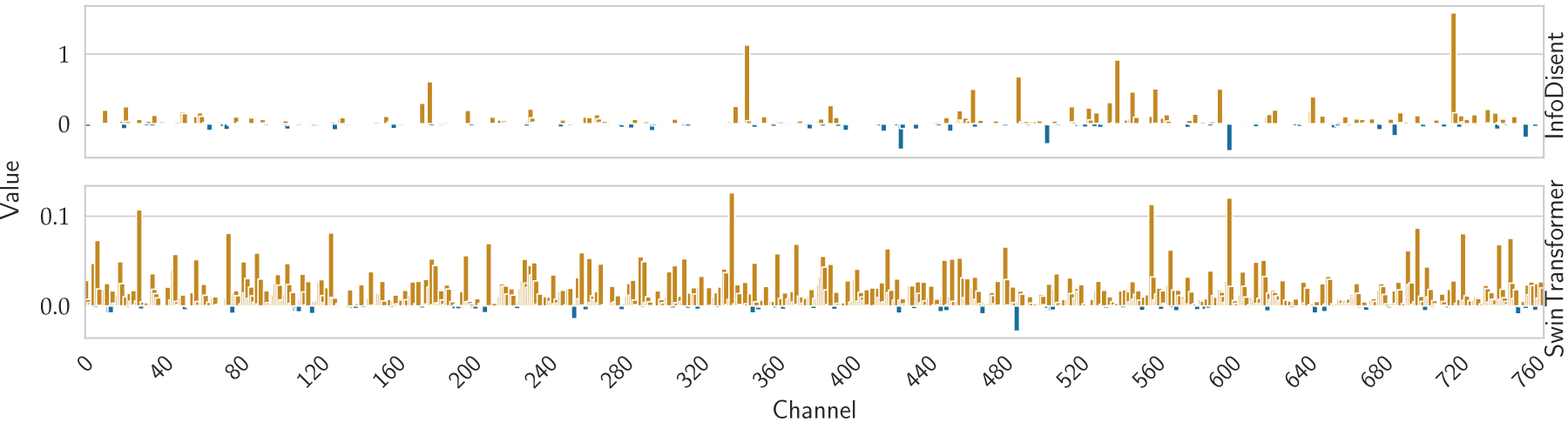}
  \caption{Channel activations before the final linear layer for a random image processed through various ConvNeXt, Vision Transformer, and Swin Transformer models trained on ImageNet. The results are displayed in three groups, each containing two graphs. Model names are listed on the right side of the graphs. \our{} activates significantly fewer channels compared to the baseline models while maintaining high effectiveness.}
  \label{fig:image_activation_channels_imagenet}
\end{figure*}

\Cref{fig:image_activation_channels_birds,fig:image_activation_channels_imagenet} also present the channel values before the final linear layer in our model for randomly selected images from various datasets and models. As evident from the images, our model utilizes a significantly smaller number of channels in its predictions compared to the baseline models.

\paragraph{Explaining Classification Decision for a Given Image by Prototypes} 
\our{} employs prototypes, similar to the approach used in PiPNet \cite{nauta2023pipnet}, to explain individual decisions made on an image. Below, we describe the process for identifying prototypes for a given input image. An average pooling operation makes the aggregation of information from individual channels. The outcome of this operation is a scalar for each channel $K$, computed as 
\[
\mathrm{mx\_pool}(K)=\max(\relu(K))-\max(\relu(-K))
\]
(as explained in the main paper). This dense representation, formed by aggregating all the channels, is then processed by a linear layer that outputs logits, see~\Cref{fig:architecture}. The logits can be represented in a format similar to the output of a convolutional layer, as illustrated by $V_{+}$ and $V_{-}$ in~\Cref{fig:architecture}. This approach maintains the pictorial structure of the logits, allowing us to extract individual channels.

\begin{figure*}[thb]
    \centering
    \includegraphics[width=\linewidth]{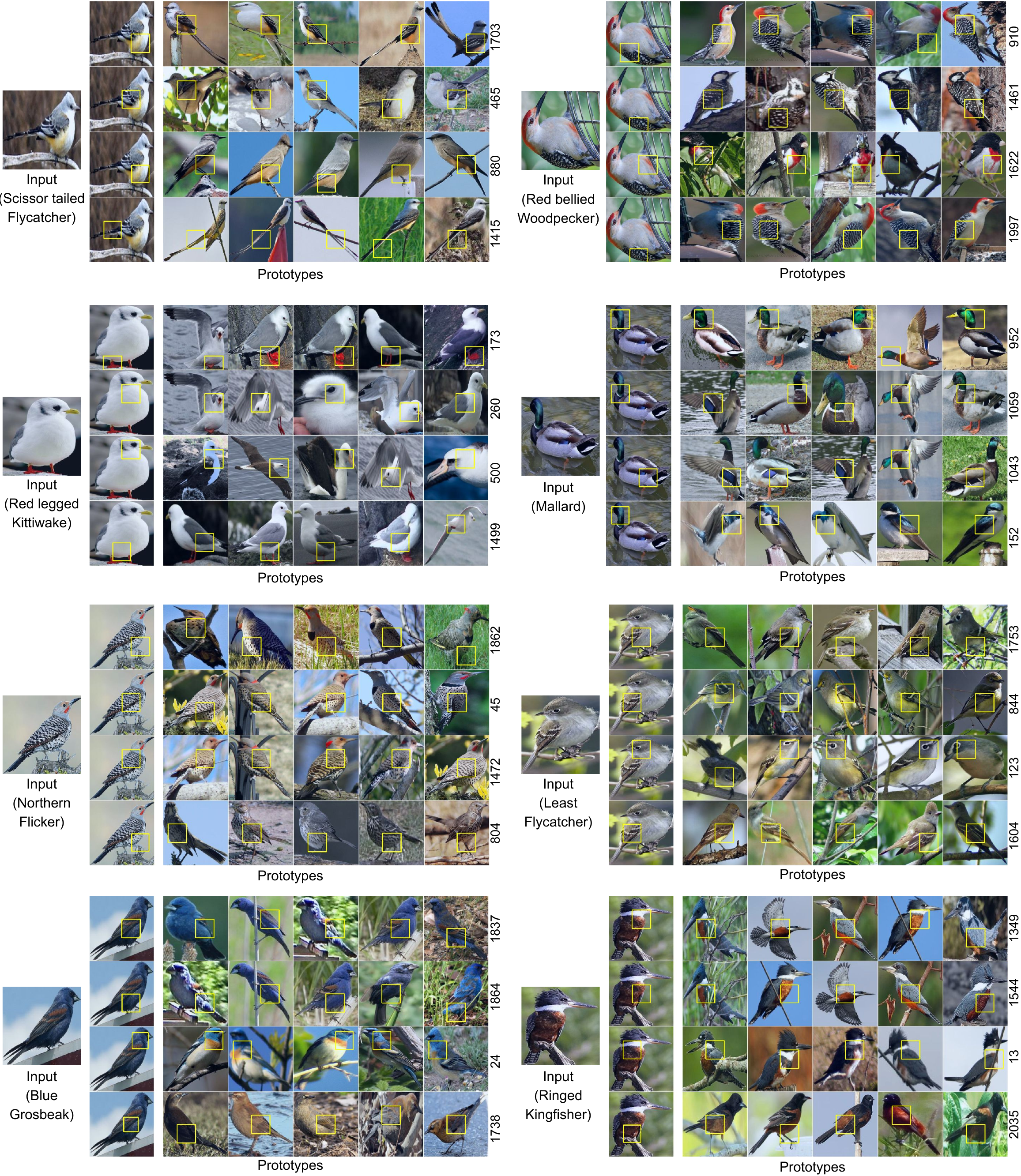}
    \caption{Example prototypes generated by \our{(ResNet-50)} models for random input images from the test set of the CUB-200-2011 dataset. The display includes 8 input images, each with a corresponding column where yellow boxes highlight specific regions, followed by the prototypes (images on the right). Each row represents prototypes from a different channel, with the channel index on the right. Observe that the prototypes identified by the model effectively capture distinct parts of the body in the images.}
    \label{fig:input_R50_birds}
\end{figure*}

\begin{figure*}[thb]
    \centering
    \includegraphics[width=\linewidth]{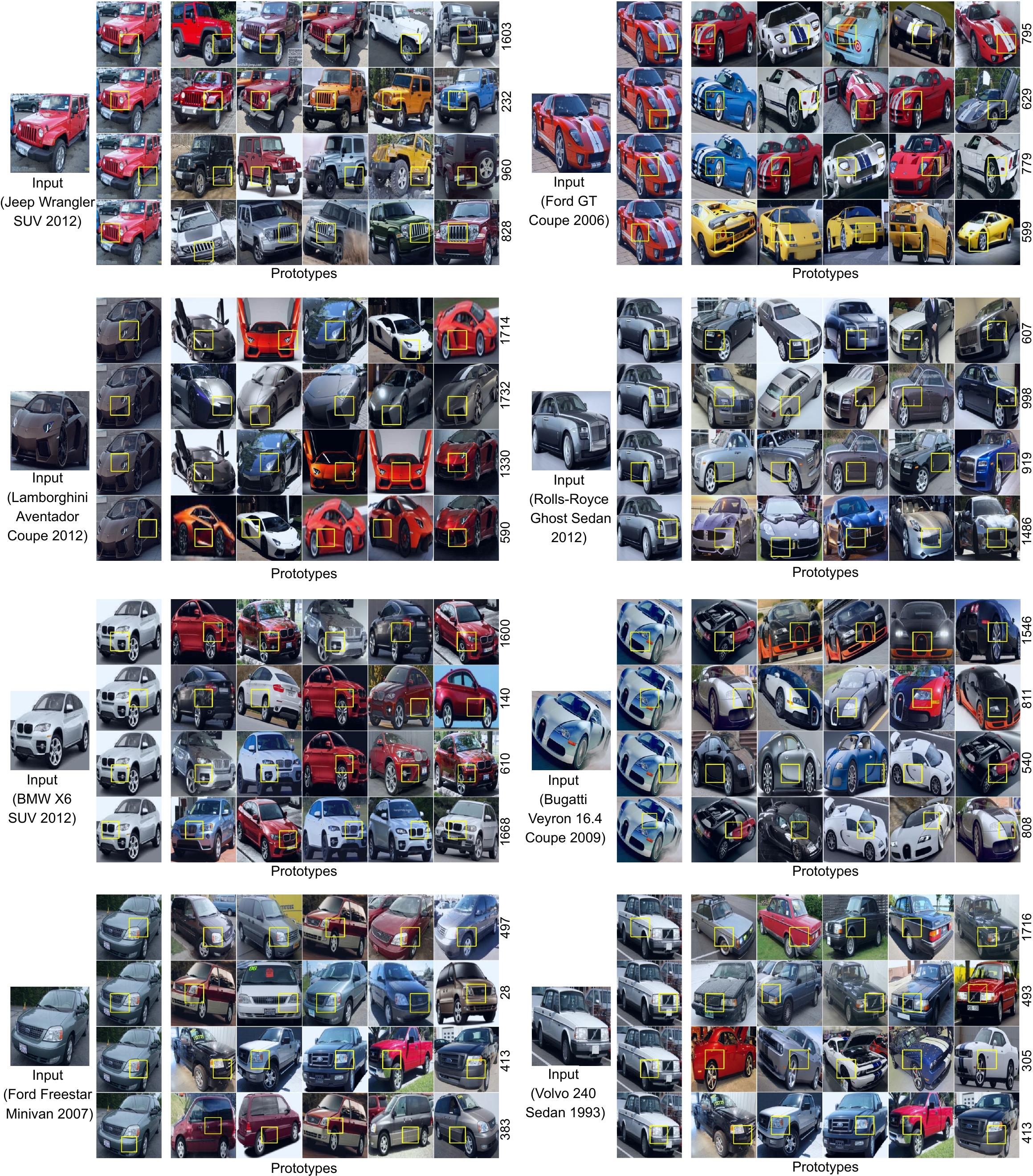}
    \caption{To demonstrate \our{(ResNet-50)} model's capability, we visualize prototypes generated for arbitrary test images from the Stanford Cars dataset. Each image is paired with highlighted areas and its associated prototypes, grouped by channel. The results indicate that the model has learned to extract meaningful features representing different vehicle parts.}
    \label{fig:input_R50_cars}
\end{figure*}

\begin{figure*}[thb]
    \centering
    \includegraphics[width=\linewidth]{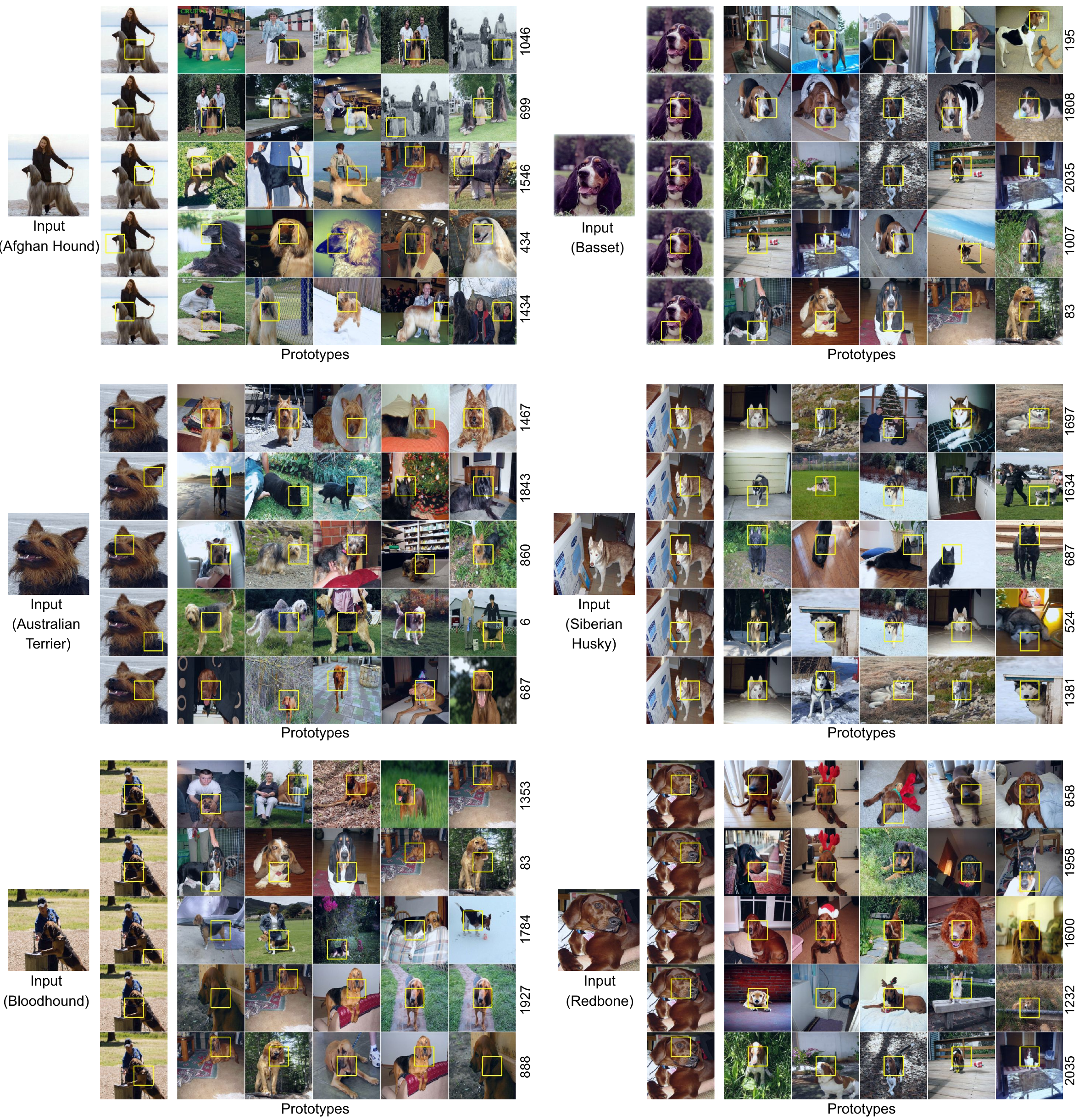}
    \caption{Visualized prototypes from the Stanford Dogs dataset. Each row highlights prototypes from a specific channel, focusing on different dog features such as ears, nose, and fur.}
    \label{fig:input_R50_dogs}
\end{figure*}

\begin{figure*}[thb]
    \centering
    \includegraphics[width=\linewidth]{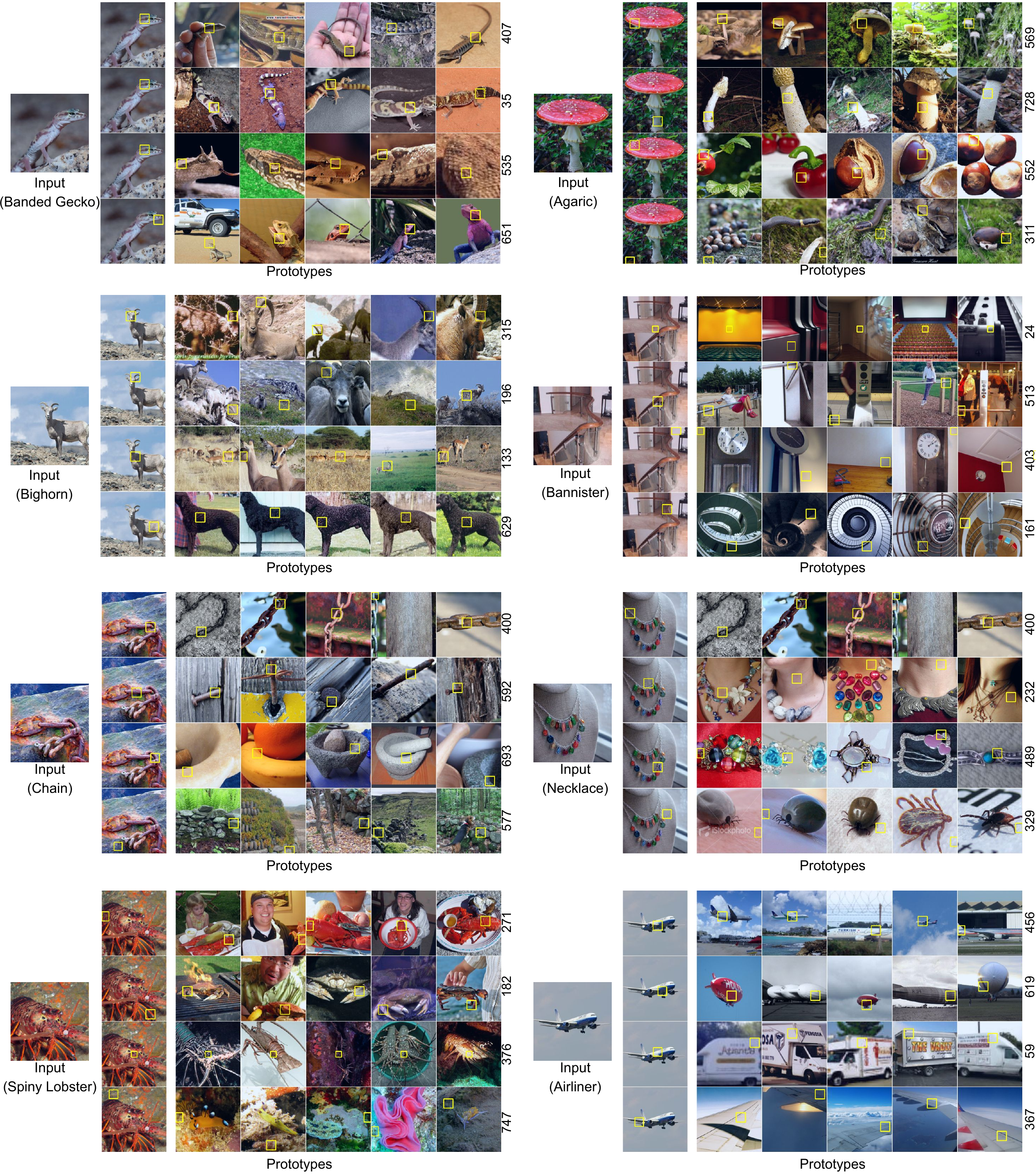}
    \caption{This figure showcases the prototypical explanations provided by \our{(ViT-B/16)} model. We visualize prototypes for arbitrary ImageNet images, highlighting relevant regions and their corresponding channel-specific prototypes.}
    \label{fig:input_prototypes_vit}
\end{figure*}

\begin{figure*}[thb]
    \centering
    \includegraphics[width=\linewidth]{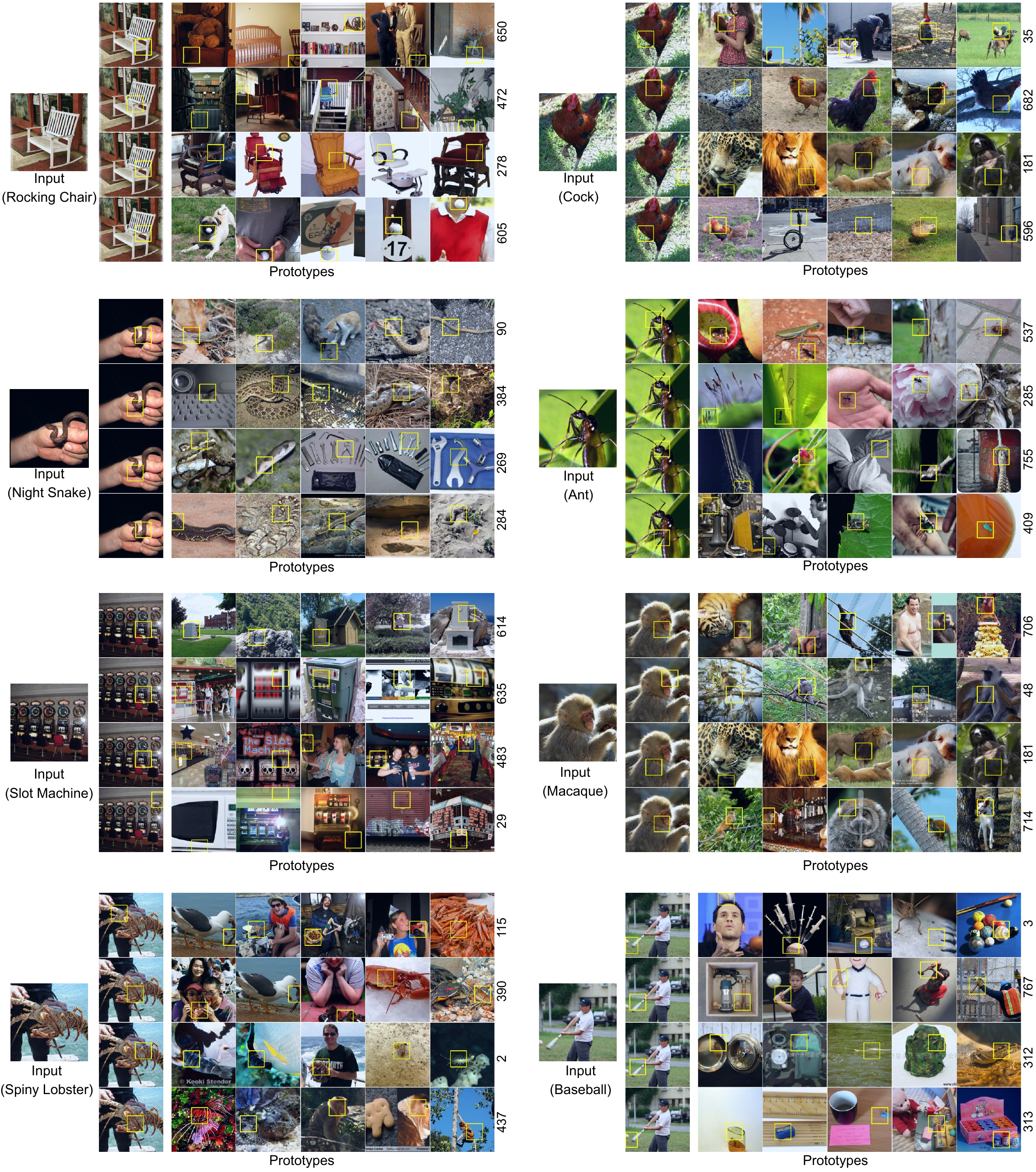}
    \caption{The figure demonstrates the prototypical explainability of \our{(Swin-S)} model by visualizing prototypes generated for arbitrary ImageNet images, highlighting relevant image regions and their corresponding channel-specific prototypes.}
    \label{fig:input_prototypes_swinT}
\end{figure*}

Note that each channel in \our{} model can contain only two possible values (refer to Figure 5a in the main paper, where these values are depicted as red or blue areas within the channels). To identify the prototype, i.e., the relevant channel, we focus solely on the positive values within the channels (represented by red areas in Figure 5a). These positive values indicate the significant part of the channel/prototype (marked by the yellow frame on the prototype) and define the channel's importance for the model prediction (since the linear layer uses only nonnegative coefficients). As demonstrated in \Cref{fig:image_activation_channels_birds,fig:image_activation_channels_imagenet}, the number of such channels is limited (alternatively, we could also focus on channels with the strongest values). Once we have identified the important channels (by knowing their indices), we represent each channel using prototypes. Prototypes are images from the training set that exhibit the five strongest positive values for a given channel. Example results illustrating how model predictions are explained using prototypes are shown in~\Cref{fig:input_R50_birds,fig:input_R50_cars,fig:input_R50_dogs,fig:input_prototypes_vit,fig:input_prototypes_swinT} (which are after the references).

\Cref{fig:input_R50_birds,fig:input_R50_cars,fig:input_R50_dogs} showcase the performance of our prototype models on standard benchmarks: CUB-200-2011, Stanford Cars, and Stanford Dogs. \Cref{fig:input_R50_birds} demonstrates the model's ability to focus on distinctive features like the Scissor-tailed Flycatcher's elongated tail feathers, underwing yellow coloration, or the Red-legged Kittiwake's red feet.

Similarly, \Cref{fig:input_R50_cars} highlights the model's capacity to identify key vehicle components. For instance, it accurately pinpoints the fender, bumper, and body of a Jeep Wrangler SUV 2012, and the characteristic stripes on a Ford GT Coupe 2006.

Complementing our experiments with widely used CNN models, we investigated the performance of transformer architectures, specifically VisionTransformer (ViT-B/16)~\cite{dosovitskiy2020image}, SwinTransformer (Swin-S)~\cite{liu2022swin}. Their results are visualized in~\Cref{fig:input_prototypes_vit,fig:input_prototypes_swinT}. Our analysis reveals that transformer models concentrate on smaller image regions than CNNs. Nevertheless, both model types generate interpretable prototypes that offer insights into the input image content.

\paragraph{Decision Behind Class}
To delve deeper into the model's decision-making process, we employ a prototype-based analysis at the class level. For each class, we identify key channels that consistently exhibit strong activation across the entire test set. These channels, indicative of the model's focus on specific visual features, are selected based on their prominence and reliability in representing the class. By visualizing these key channels as prototypes, as demonstrated in our previous image analysis, we obtain a clear representation of the model's class-specific decision criteria.

\begin{figure*}[thb]
    \centering
    \includegraphics[width=\linewidth]{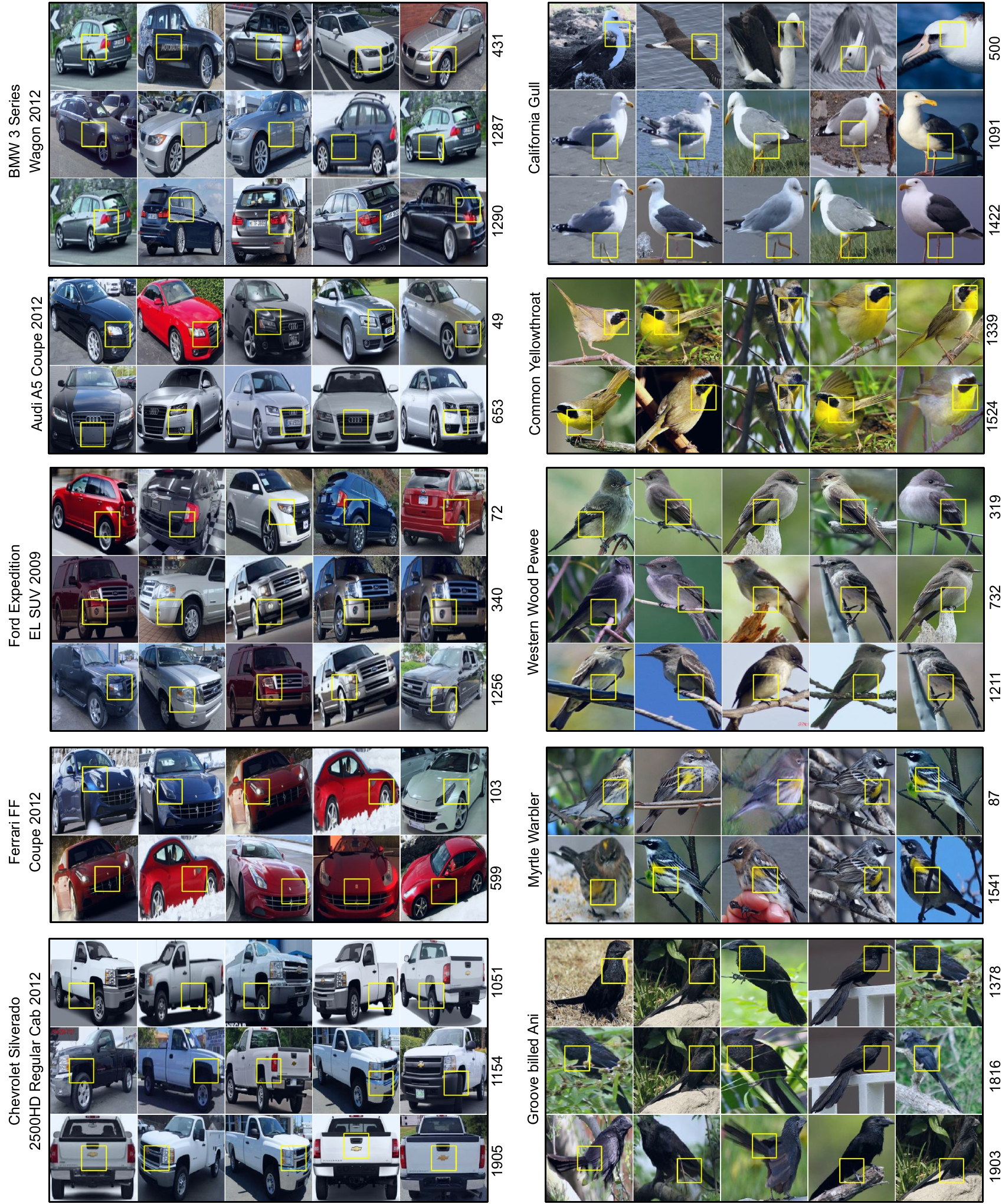}
    \caption{Prototypes of sample classes from \our{(ResNet-50)} model trained on the cropped Stanford Cars dataset (left) and the CUB-200-2011 dataset (right). Yellow frames highlight the class prototypes.}
    \label{fig:prototyps_class_R50}
\end{figure*}

\begin{figure*}[thb]
    \centering
    \includegraphics[width=.99\linewidth]{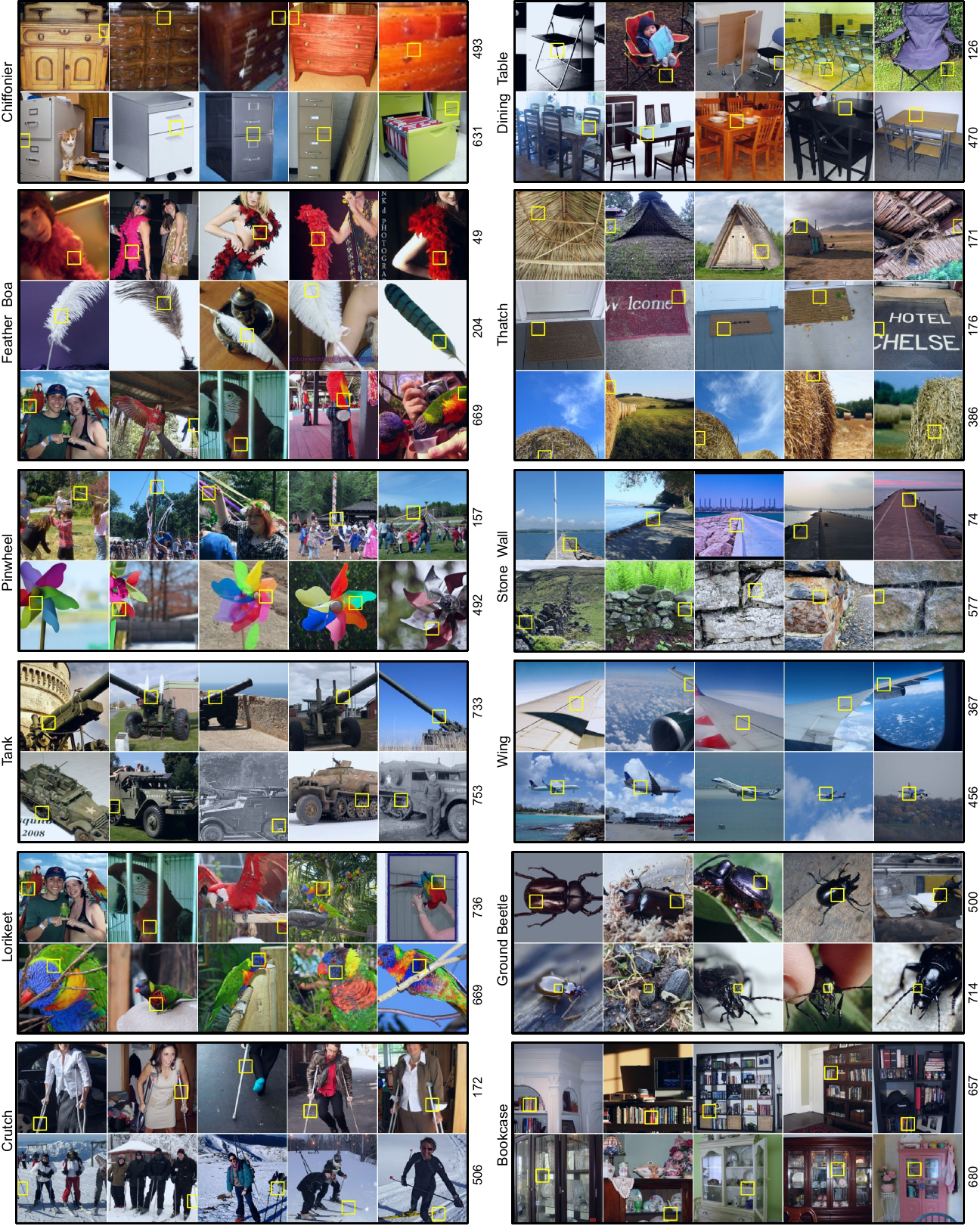}
    \caption{Visualizations of \our{(ViT-B/16)} model-generated prototypes for selected ImageNet categories.}
    \label{fig:prototyps_class_vit}
\end{figure*}

\begin{figure*}[thb]
    \centering
    \includegraphics[width=\linewidth]{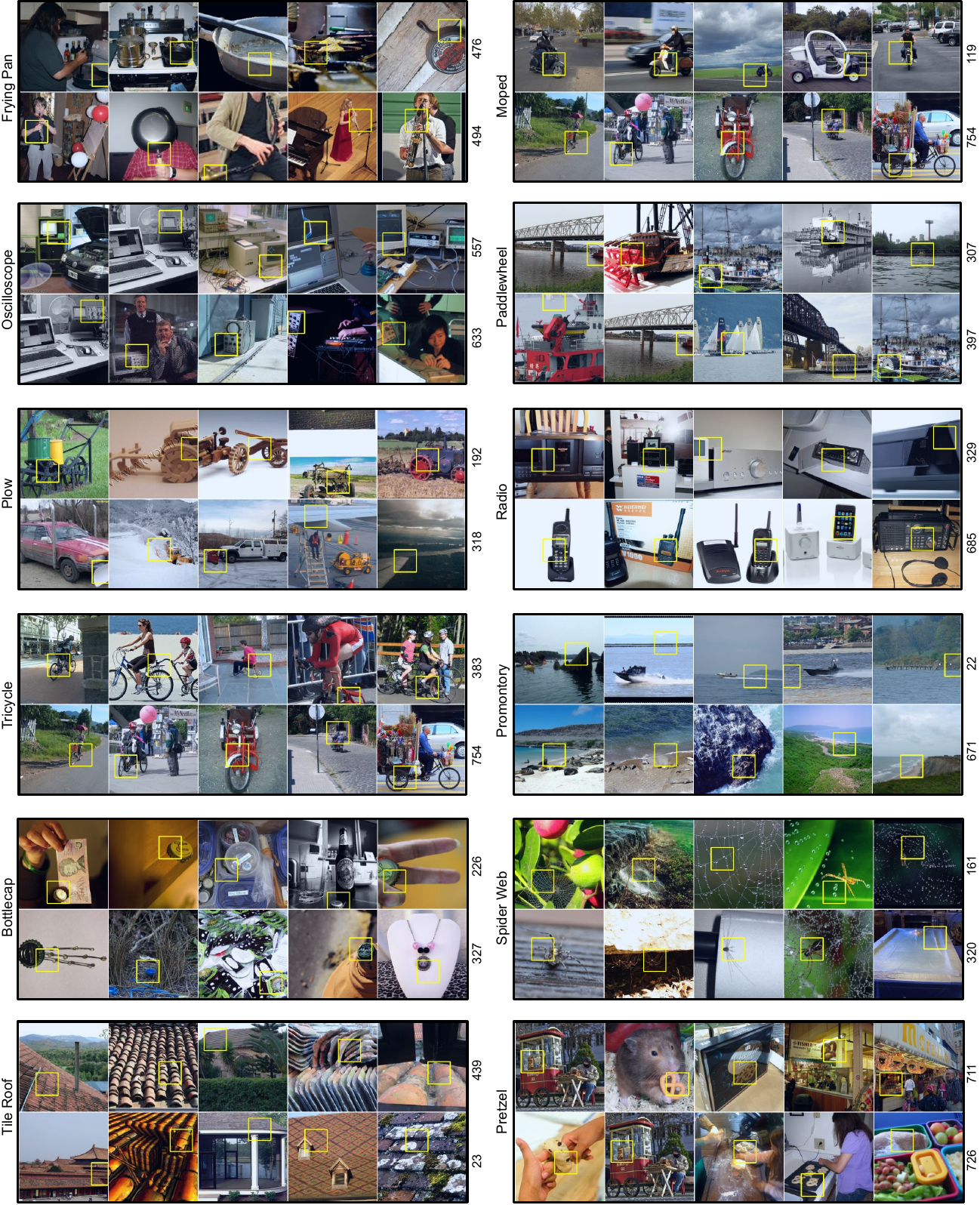}
    \caption{Representative ImageNet class prototypes produced by the \our{(Swin-S)} model.}
    \label{fig:prototyps_class_swinT}
\end{figure*}

\Cref{fig:prototyps_class_R50,fig:prototyps_class_vit,fig:prototyps_class_swinT} present the results of our prototype analysis for selected classes in the CUB-200-2011 and Stanford Cars datasets, using both \our{(ResNet-50)} and transformer models. This visualization allows for a granular understanding of how these models differentiate between various classes, highlighting the underlying patterns recognized by the network.

\paragraph{Heatmaps}
\Cref{fig:heatmaps,fig:heatmaps_imagenet} present example heatmaps generated by our model, resembling those produced by Grad-CAM~\cite{selvaraju2017grad}. Our method, rooted in representational channels, simplifies heatmap generation by accumulating activations from all prototypes (positive and negative) across all channels. Leveraging the information bottleneck principle, our approach yields more focused heatmaps compared to traditional methods.

\paragraph{More details on user study}

Each worker was paid €2.00 for completing a short 20-question survey. The survey questions were randomly composed, so the specific questions differed between participants. The participants were gender-balanced and ranged in age from 18 to 60. They were given 30 minutes to complete the survey.

To ensure data quality, we excluded responses where users selected the same answer for all questions. Surveys were repeated until we obtained 60 valid responses. Figure~\ref{fig:us1} and Figure~\ref{fig:us2} illustrate example questions used in both user studies.

Before starting the survey, participants were provided with an example and detailed instructions to familiarize them with the study setup, including the explanation composition and visualization. The distribution of answers is summarized in Tables~\ref{tab:us1} and Table~\ref{tab:us2}.

\begin{table*}[t]\small
    \centering
    \caption{Distribution of answers in user study on user confidence in model's prediction based on explanation.}
    \begin{tabular}{@{}c@{}c@{}c@{}c@{}c@{}}
    \toprule
        Dataset & Fairly ... correct & Somewhat ... correct & Somewhat ... incorrect & Fairly ... incorrect \\
    \midrule
        CUB & 585 & 316 & 208 & 91\\
        ImageNet & 449 & 248 & 189 & 314\\
    \bottomrule
    \end{tabular}
    \label{tab:us1}
\end{table*}

\begin{table}[t]\small
    \centering
    \caption{Distribution of answers in user study on prototypical part disambiguity.}
    \begin{tabular}{@{}c@{\qquad}cc@{}}
    \toprule
        Dataset & Correct & Incorrect \\
    \midrule
        CUB & 776 & 424  \\
        ImageNet & 712 & 488  \\
    \bottomrule
    \end{tabular}
    \label{tab:us2}
\end{table}

\paragraph{Is \our{} reducing channel correlation?}  To assess whether our model reduces channel correlation, we computed the RV coefficient, a standard measure of linear correlation between vectors. In 14 experiments (across 3 datasets and 7 backbones), our model outperformed the baseline in 9 cases (smaller RV coefficients), performed worse in 2 cases, and showed equivalent results in 3 cases as presented in Table~\ref{tab:rv}. Despite not having a loss term explicitly aimed at reducing the RV coefficient, our model generally demonstrates either smaller or equal RV coefficients compared to the baseline.

\begin{table}[t]\small
    \centering
    \caption{RV correlation between channels for \our{} and baseline models.}
    \renewcommand{\arraystretch}{1.2} 
    \begin{tabular}{@{}lccc@{}}
        \toprule
        \multicolumn{1}{c}{Dataset} & Model & \our{} & Baseline \\
        \midrule
        \multirow{4}{*}{Cars} 
        & ConvNeXt Tiny & \textbf{6.7} & 6.9 \\
        & DenseNet121 & \textbf{6.2} & 6.9 \\
        & ResNet34 & \textbf{6.0} & 7.1 \\
        & ResNet50 & \textbf{6.2} & 7.6 \\
        \midrule
        \multirow{4}{*}{CUB-200-2011} 
        & ConvNeXt Tiny & \textbf{5.5} & \textbf{5.5} \\
        & DenseNet121 & \textbf{7.2} & 7.9 \\
        & ResNet34 & \textbf{6.5} & 7.2 \\
        & ResNet50 & \textbf{4.9} & 7.8 \\
        \midrule
        \multirow{6}{*}{ImageNet} 
        & ConvNeXt Large & 8.7 & \textbf{3.0} \\
        & DenseNet121 & 8.4 & \textbf{4.5} \\
        & ResNet34 & \textbf{5.2} & 6.0 \\
        & ResNet50 & \textbf{3.0} & 6.5 \\
        & SwinV2-S & \textbf{2.5} & \textbf{2.5} \\
        & ViT-B-16 & \textbf{9.9} & \textbf{9.9} \\
        \bottomrule
    \end{tabular}
    \label{tab:rv}
\end{table}

\begin{figure*}[thb]
    \centering
    \includegraphics[width=.85\linewidth]{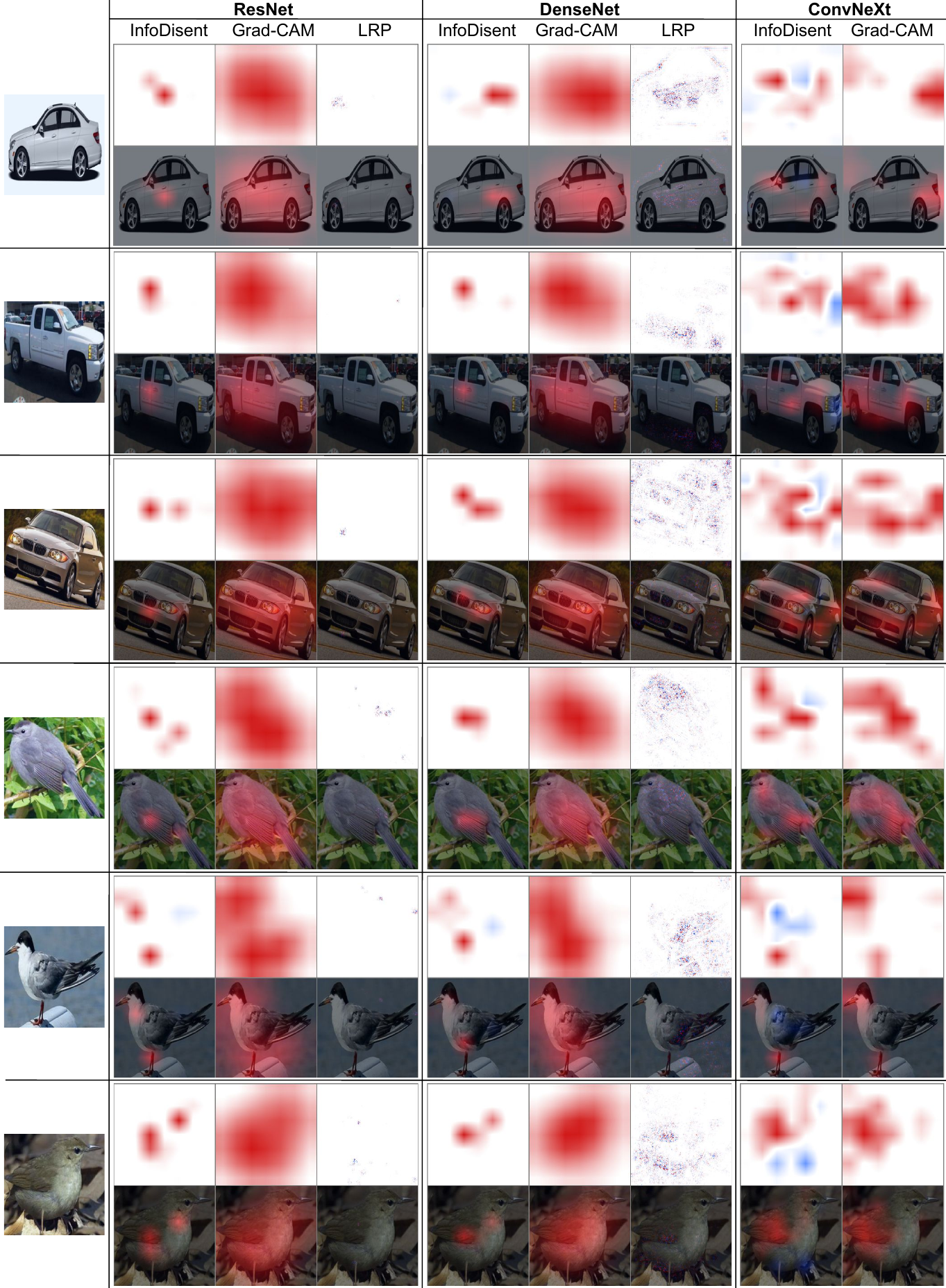}
    \caption{Comparison of example heat maps generated by \our{} proposed model and competing approaches. Our technique produces more focused heat maps by leveraging representational channels and the information bottleneck principle, outperforming traditional methods like Grad-CAM.}
    \label{fig:heatmaps}
\end{figure*}

\begin{figure*}[thb]
    \centering
    \includegraphics[width=.8\linewidth]{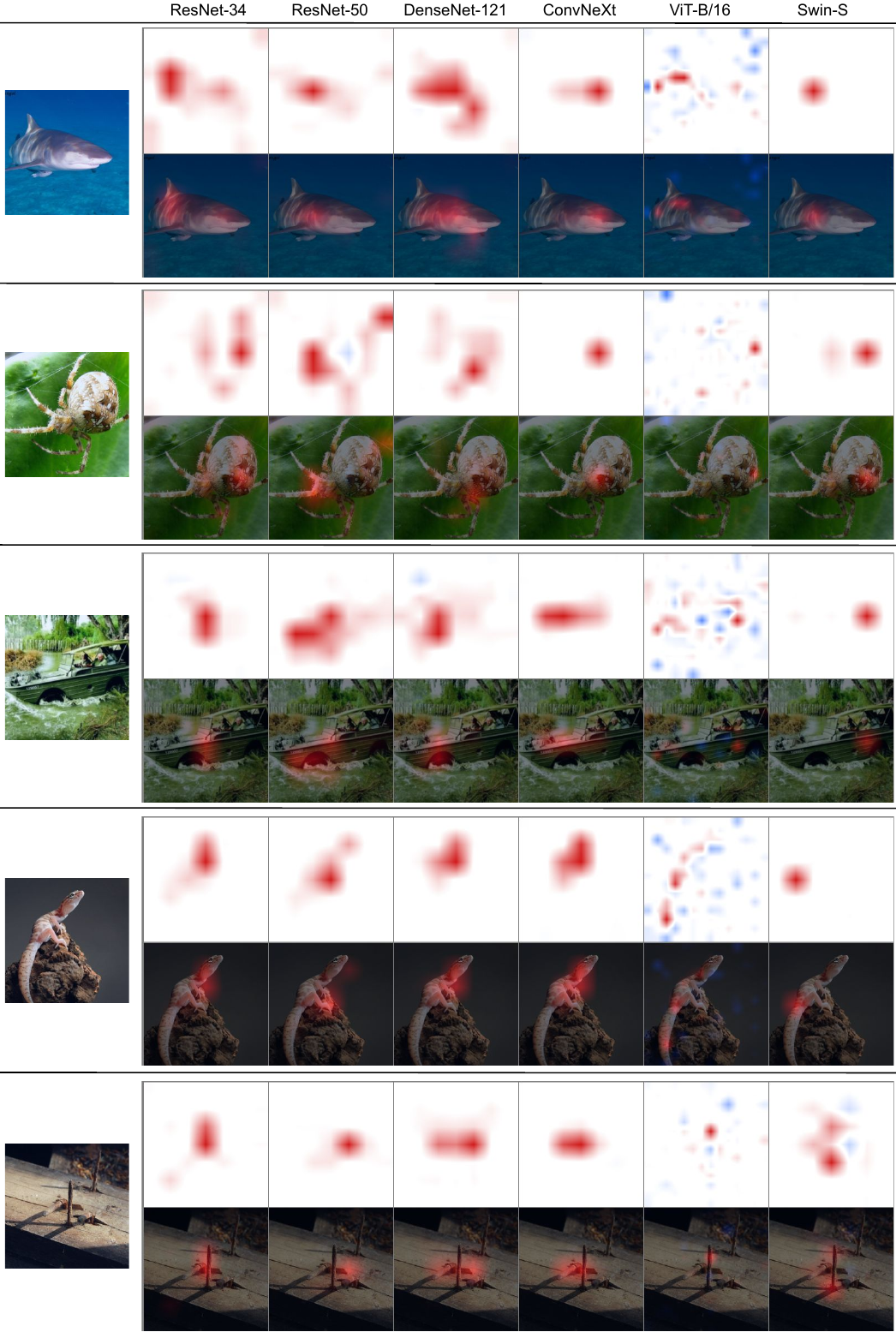}
    \caption{Example heatmaps generated by \our{}, our proposed method, demonstrating activation regions on sample photos from the ImageNet test set.}
    \label{fig:heatmaps_imagenet}
\end{figure*}

\begin{figure*}
    \centering
    \includegraphics[width=0.6\textwidth]{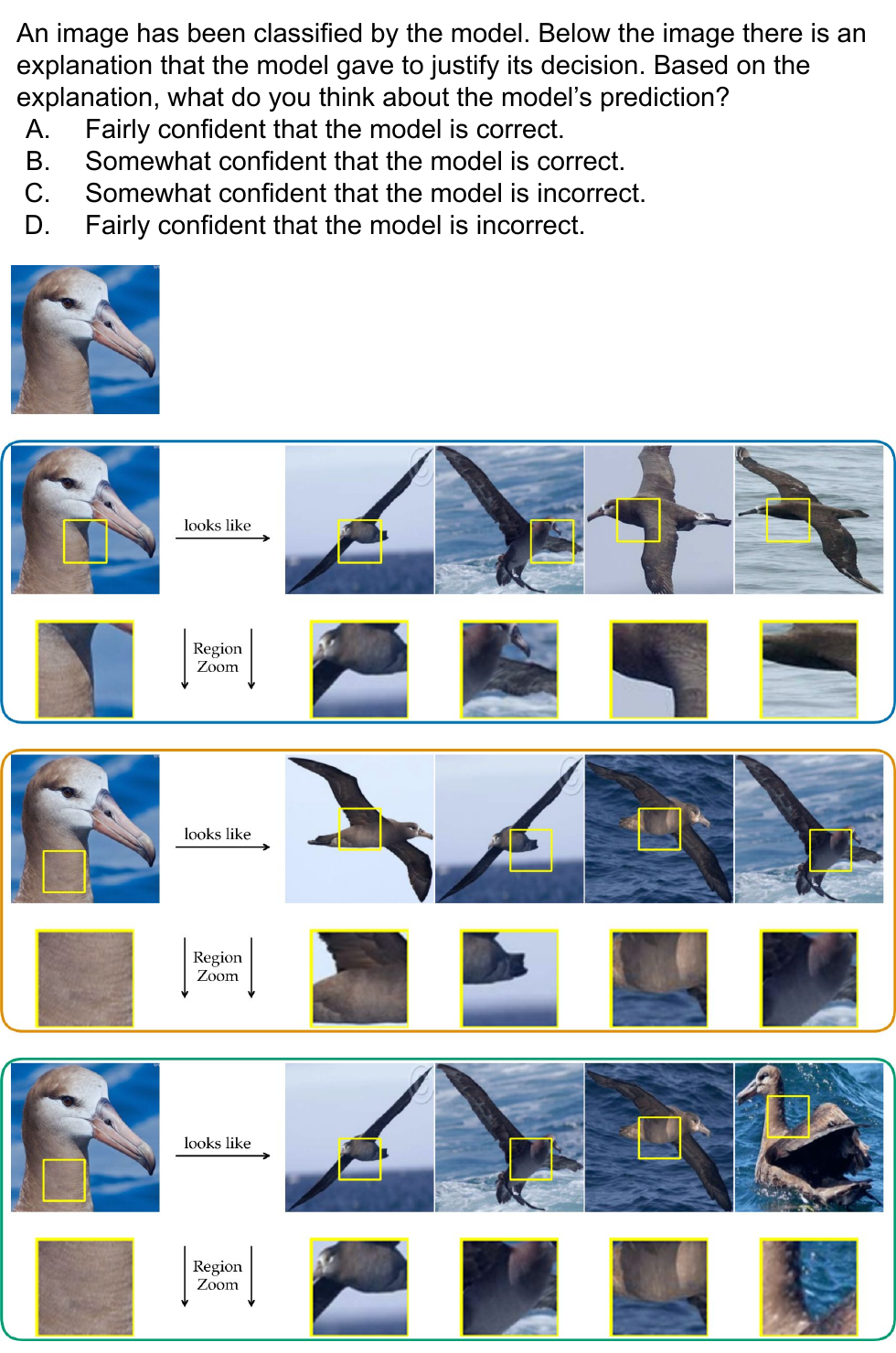}
    \caption{An exemplary question from the user study on user confidence.}
    \label{fig:us1}
\end{figure*}

\begin{figure*}
    \centering
    \includegraphics[width=\textwidth]{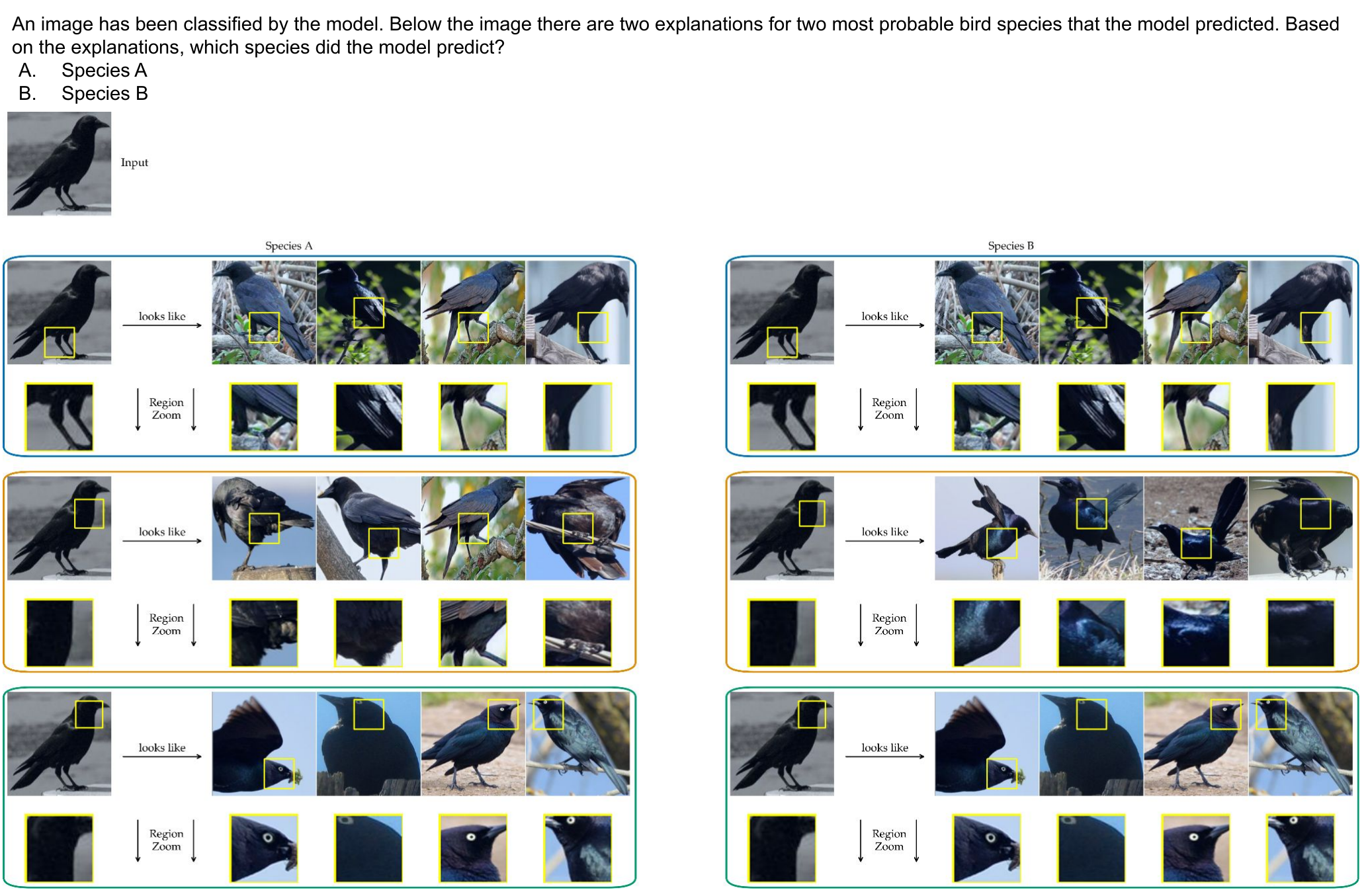}
    \caption{An exemplary question from the user study on disambiguity of prototypical parts.}
    \label{fig:us2}
\end{figure*}

\clearpage

\end{document}